\documentclass{postunc}
\usepackage{tikz}
\usepackage{amsmath}
\usepackage{cleveref}
\usepackage{parskip}
\usepackage[square,numbers,sort&compress]{natbib}
\usepackage[charter]{mathdesign}
\usepackage{microtype}
\usepackage{xcolor}

\title{Meta-models for transfer learning in source localisation}

\author{Lawrence A.\ Bull$^{1,2*}$, Matthew R.\ Jones$^{3*}$, Elizabeth J.\ Cross$^{3}$, Andrew Duncan$^{4,5}$, Mark~Girolami$^{2,5}$}

\heading{L.A.\ Bull et al.}

\address{
  $^1$
  University of Glasgow, School of Mathematics and Statistics, Glasgow, G12 8TA, Scotland\\
  e-mail: lawrence.bull@glasgow.ac.uk \and
  $^2$University of Cambridge, Department of Engineering, Cambridge, CB3 0FA, UK\and
  $^3$
  University of Sheffield, Department of Engineering, S1 3JD, UK\and
  $^4$
  Imperial College London, Department of Mathematics, London, SW7 2AZ, UK\and
  $^5$
  The Alan Turing Institute, The British Library, London, NW1 2DB, UK\and
  $*$
  These authors contributed equally}


\abstract{
In practice, non-destructive testing (NDT) procedures tend to consider experiments (and their respective models) as distinct, conducted in isolation and associated with independent data. %
In contrast, this work looks to capture the interdependencies between acoustic emission (AE) experiments (as meta-models) and then use the resulting functions to predict the model hyperparameters for previously unobserved systems. %
We utilise a Bayesian multilevel approach (similar to deep Gaussian Processes) where a higher level \textit{meta-model} captures the inter-task relationships. %
Our key contribution is how knowledge of the experimental campaign can be encoded \textit{between} tasks as well as within tasks. %
We present an example of AE time-of-arrival mapping for source localisation, to illustrate how multilevel models naturally lend themselves to representing aggregate systems in engineering. %
We constrain the meta-model based on domain knowledge, then use the inter-task functions for transfer learning, predicting hyperparameters for models of previously unobserved experiments (for a specific design).}

\begin{document}

\maketitle

\section{Introduction: Multilevel models for meta-modelling}

Increasingly, engineering systems are equipped with sensors, often providing streams of telemetry data. %
As the number of instrumented systems grows, \textit{population} data become available~\cite{gardner2022pbshm}. %
While machine populations clearly differ from the natural examples in ecology, epidemiology, and behavioural science, the same statistical methods can be used to represent artificial systems~\cite{Rahwan2019MB}. %

In this article, we consider \textit{multilevel} (or hierarchical) models~\cite{bda3} which are particularly suited to population data, as they naturally exhibit a hierarchical structure. %
As an engineering example, consider turbines of the same specification in a wind farm. %
Each individual is unique, with its own wind-power relationship, depending on the local environment. %
On the other hand, some parameters are shared between operating subgroups (e.g.\ maximum power) while others are global (e.g.\ the minimum rpm to generate power).

A multilevel model can represent interdependent population data via \textit{partial pooling}~\cite{bda3}
where parameter hierarchies are learnt with machine-specific and shared parameters (or functions). %
As such, models share (i.e.\ pool) information, to extend the value of data.
Multilevel models are often termed \textit{multitask learners}, as multiple related tasks $f_k$ are learnt simultaneously to improve inference \cite{murphy2012machine}. %
\Cref{fig:visualise} visualises how multilevel models can learn multiple tasks. %

\begin{figure}[h]
    \centering
    \resizebox{.99\textwidth}{!}{%
    \begin{tikzpicture}
        \linespread{1}
        \tikzstyle{block} = [rectangle, thick, draw, rounded corners, text width=3em, text centered, minimum height=5em, fill=black!40, node distance=8em]
        
        \node [block, fill=purple!40] (f1) {$f_1$};
        \node [right of=f1, block, fill=purple!40] (f2) {$f_2$};
        \node [right of=f2, block, fill=purple!40] (f3) {$f_3$};
        \node [right of=f3, xshift=3em] (f4) {$\ldots$};
        
        \node [circle, above of=f4, fill=black!30, node distance=12em] (theta) {$\;\; g^\prime \rightarrow \theta^\prime_k \;\;$};
        
        \node [right of=f4, block, fill=purple!40] (f5) {$f_{K-2}$};
        \node [right of=f5, block, fill=purple!40] (f6) {$f_{K-1}$};
        \node [right of=f6, block, fill=purple!40] (f7) {$f_K$};
        
        \node [circle, above of=f2, fill=black!15, node distance=12em] (theta1) {$g^{\prime\prime} \rightarrow \theta^{\prime\prime}_k$ };
        \node [circle, above of=f6, fill=black!15, node distance=12em] (theta2) {$g^{\prime\prime\prime} \rightarrow \theta^{\prime\prime\prime}_k$};
        
        
        \path [draw, thin] (theta) -- node[fill=white,pos=.3] {$\theta^\prime_1$} (f1);
        \path [draw, thin] (theta) -- node[fill=white,pos=.3] {$\theta^\prime_2$} (f2);
        \path [draw, thin] (theta) -- node[fill=white,pos=.3] {$\theta^\prime_3$} (f3);
        \path [draw, thin] (theta) -- node[fill=white,pos=.3] {$\theta^\prime_{k-2}$} (f5);
        \path [draw, thin] (theta) -- node[fill=white,pos=.3] {$\theta^\prime_{k-1}$} (f6);
        \path [draw, thin] (theta) -- node[fill=white,pos=.3] {$\theta^\prime_{K}$} (f7);
        
        \path [draw, thin] (theta1) -- node[fill=white,pos=.6] {$\theta^{\prime\prime}_1$} (f1);
        \path [draw, thin] (theta1) -- node[fill=white,pos=.6] {$\theta^{\prime\prime}_2$} (f2);
        \path [draw, thin] (theta1) -- node[fill=white,pos=.6] {$\theta^{\prime\prime}_3$} (f3);
        
        \path [draw, thin] (theta2) -- node[fill=white,pos=.6] {$\theta^{\prime\prime\prime}_{k-2}$} (f5);
        \path [draw, thin] (theta2) -- node[fill=white,pos=.6] {$\theta^{\prime\prime\prime}_{k-1}$} (f6);
        \path [draw, thin] (theta2) -- node[fill=white,pos=.6] {$\theta^{\prime\prime\prime}_{K}$} (f7);
        \end{tikzpicture}}
    \caption{A visual example of multilevel models for multitask learning: predictive tasks $f_k$ are parametrised by $\theta_k = \{\theta^\prime_k, \theta^{\prime\prime}_k\}$ or $\theta_k =\{\theta^\prime_k, \theta^{\prime\prime\prime}_k\}$; in turn, $\theta_k$ is generated by shared intertask functions $\{g^\prime, g^{\prime\prime}\}$ or $\{g^\prime, g^{\prime\prime\prime}\}$. The notation $g \rightarrow \theta$ shows a function $g$ that predicts a parameter $\theta$.}
    \label{fig:visualise}
\end{figure}

With interpretable models, one can inspect how parameters $\theta_k$ vary between tasks. %
These variations are termed \textit{intertask relationships}, and each function that approximates them (shown as $g$ in \Cref{fig:visualise}) can be considered as a \textit{meta-model} (a \textit{model of models}). %
If desired, these intertask functions can be built to vary with respect to higher-level (\textit{macro}) explanatory variables. %
A simple example is the varying coefficients model, typically demonstrated with the 8-schools data \cite{kreft1998introducing}. %
Intertask relationships are useful in engineering practice, as they allow alternative insights to be extracted from collected population data. %

\subsection{Source localisation}
This work considers a population where members are represented by a sequence of 28 Acoustic Emission (AE) experiments, originally from~\cite{hensman2010locating}. %
In each experiment, the arrival time of waves propagating through a complex plate geometry is recorded. %
For each dataset, a 2D map of the arrival times is learnt through regression. %
Despite distinct experiment designs, the metal plate remains consistent (the medium through which waves propagate) suggesting that their models share certain parameters. %
By analysing these datasets collectively, as a population, additional insights are extracted from the test campaign. %
We capture how \textit{characteristics} of the 2D map vary, allowing predictions of hyperparameters for the response surface of experimental designs that were absent from the training data. %
In other words, one can extrapolate or interpolate in the \textit{model space} by using the intertask functions as meta-models, to predict hyperparameters. %
Because of the plate's complexity, we encode weak constraints on the model given domain expertise of the experimental campaign, rather than specific physics-based laws (e.g. via differential equations). %

These insights have significant implications since the meta-models can be used to predict the characteristics of the response (hyperparameters) for new, previously unobserved experimental designs. %
Such \textit{model} predictions alleviate the requirement of extensive training data in new tasks, by sharing information from similar experiments. %
The result is an interpretable and explainable approach to transfer learning for engineering experiments. %

\subsection{Layout}

\Cref{s:AE} introduces the AE data, the experimental campaign, and their multilevel interpretation. %
\Cref{s:stl} uses Gaussian Process regression for a single AE experiment. %
\Cref{s:mtl} extends the model to represent data from the full test campaign in a joint inference, investigating different model assumptions and the associated intertask relationships. %
\Cref{s:res} assesses the predictive performance and utilises the model for transfer learning. %
\Cref{s:conc} offers concluding remarks. %

\subsection{Related Work \& Contribution}
To improve interpretability and ensure meaningful outputs from data-driven models, the inclusion of physical insight within machine learning methods is becoming increasingly popular, with overviews found in~\cite{karniadakis2021physics,willard2020integrating}. %
These approaches are often referred to as physics-informed machine learning, since they utilise both data and explanatory physics to improve modelling when compared to either approach independently. %
Some examples include physics-guided loss functions~\cite{daw2021physicsguided}, vector field constraints~\cite{wahlstrom2013modeling}, and the inclusion of governing differential equations~\cite{alvarez2009latent}. %
Given the complexity of the plate geometry in this work, the physics-based constraints are \textit{soft}, since any predictive functions can deviate from the suggested structure. %
To achieve this, a Bayesian approach is used, where the constraints are placed on a (hierarchical) prior distribution \cite{bda3}, which offers a natural trade-off between prior knowledge and data. %
While this trade-off is possible with a (weighted) physics-guided loss function, we favour a hierarchical statistical approach for uncertainty quantification and interpretability when multitask learning. %

\paragraph{Contribution}
In the context of AE (time of arrival) mapping, previous work of (coauthors) \citet{jones2023constraining} considered how domain knowledge can be encoded in Gaussian Process (GP) models via boundary condition constraints, applied to the kernel function for a single experiment. %
Here, we extend the work to represent multiple experiments, allowing us to encode domain expertise at the \textit{systems level} and infer intertask relationships for a series of tests
\footnote{In this work, we use the term \textit{systems level} to refer to the combined tasks, which allow parameters or functions to be learnt between multiple experiments; i.e. the inter-task functions.}. %
Our key contribution is the ability to encode physics-based knowledge \textit{between} tasks, as well as within tasks. %
Rather than a single experiment, the resultant model represents variations over collected tests in an experimental campaign: this allows for simulation and (hyper) parameter prediction at previously unobserved experimental designs (in this case, sensor separation). %

A multilevel modelling approach is adopted, which is increasingly utilised in the engineering literature. %
An early monitoring example is presented by \citet{huang2019multitask} and \citet{huang2015hierarchical}, where multiple correlated regression tasks are utilised for modal analysis. %
A shared sparsity profile is inferred for tasks relating to measurement channels to improve damage detection by considering the correlation between damage scenarios or adjacent sensors. %
More recent applications include \citet{di2021decision} who use multilevel models to represent corrosion progression given evidence from multiple locations, and \citet{papadimas2021hierarchical}, where the results from materials tests (i.e.\ coupon samples) are combined to inform the estimation of material properties. %
Similar to \citet{papadimas2021hierarchical} our work considers an experimental campaign, but rather than infer the higher-level representation as a global estimate of a single experiment, we introduce inter-task explanatory variables. %
In turn, the model represents task variations as functions, rather than uni-modal sampling distributions.

On a related theme, \citet{hughes2023towards} recognise structures and their populations as nested hierarchies, proposing a convenient formulation for decision analyses. %
\citet{sedehi2023integration} also present work to encoding physics into hierarchical GPs, where time-history measurements are partitioned into multiple segments to create longitudinal data, accounting for temporal variability and addressing the non-stationarity of the measured responses for a single structure.

\section{The Acoustic Emission Experiments}\label{s:AE}

Acoustic Emissions (AE) are ultrasonic signals released within a material as its internal structure undergoes some irreversible change. %
The driving mechanisms often relate to the initiation and growth of damage, so monitoring AE signals can serve to assess the condition of materials and structures~\cite{shull2002nondestructive}.
As emissions propagate through a material from the point of origin, differences in the time of arrival at separate sensors (in an array) can be used for \textit{triangulation} (or \textit{trilateration}) to enable source localisation~\cite{tobias1976acoustic}. %
From a damage monitoring perspective, source localisation provides an operator with more insight to make better maintenance and planning decisions.

\begin{figure}[h]
\centering
\scalebox{1}[-1]{\includegraphics[width=.5\linewidth]{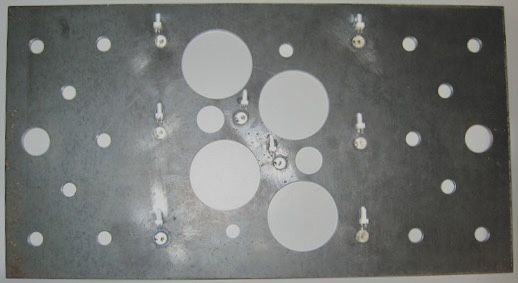}}
\caption{Image of plate used in the AE experiments.}
\label{fig:real-plate}
\end{figure}

One strategy for localising AE signals involves learning the map of the arrival times across the surface of interest~\cite{jones2022bayesian}. %
That is, the forward mapping from AE source localisation $\mathbf{x}_i$ to the measured difference in time-of-arrival ($\Delta$ToA) for a given sensor pair,
\begin{align}
y_i &= f(\mathbf{x}_i) + \epsilon_i 
\end{align}

\noindent where $y_i$ is some \textit{noisy} observation of $\Delta$ToA with additive observation noise $\epsilon_i$. %
In this paper, we consider data from experiments by Hensman et al.~\cite{hensman2010locating} concerning an aluminium plate-like structure shown in \Cref{fig:real-plate}. 

\begin{figure}[ht]
\centering
\includegraphics[width=.46\linewidth,valign=b]{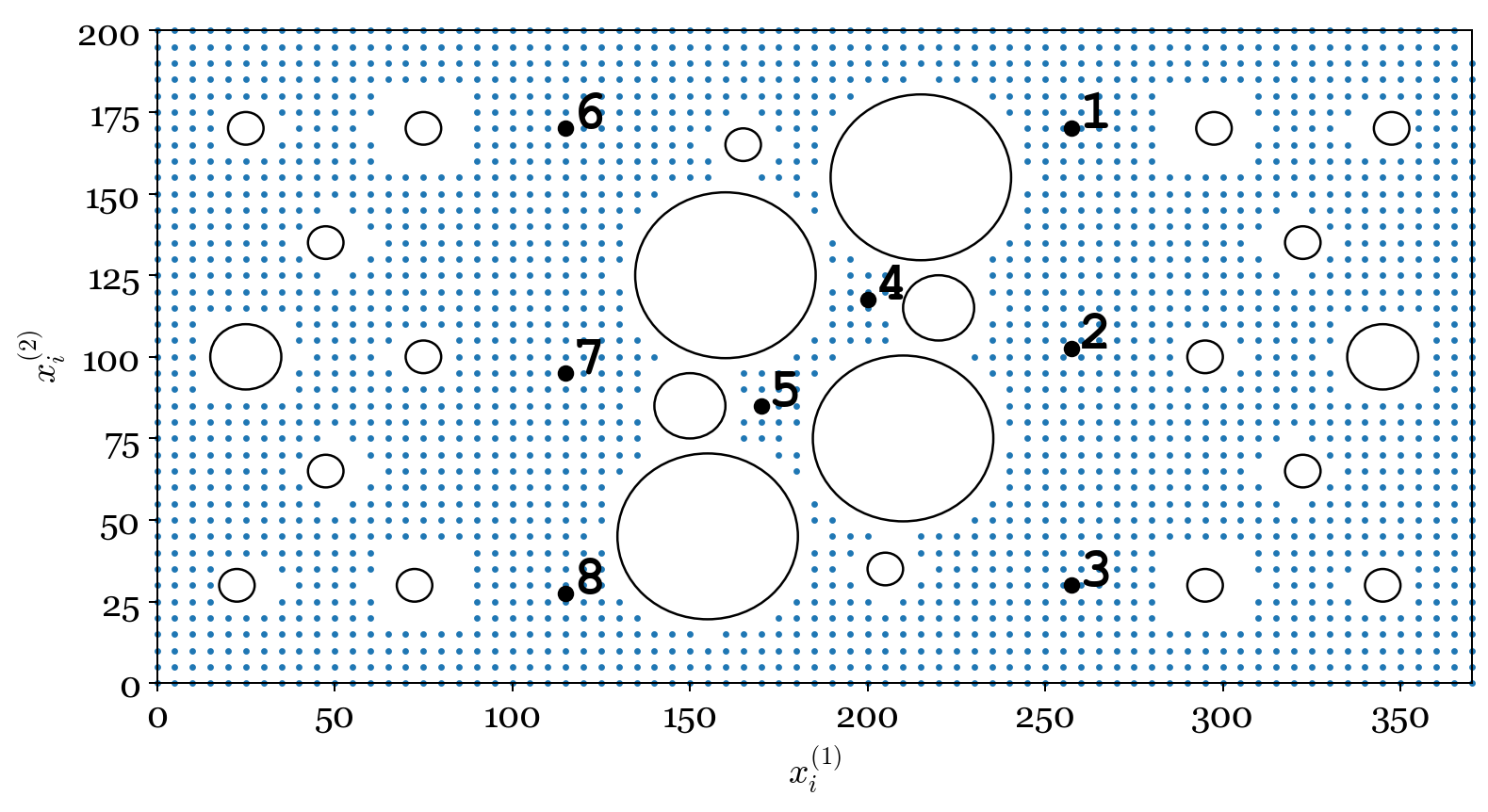}%
\includegraphics[width=.54\linewidth,valign=b]{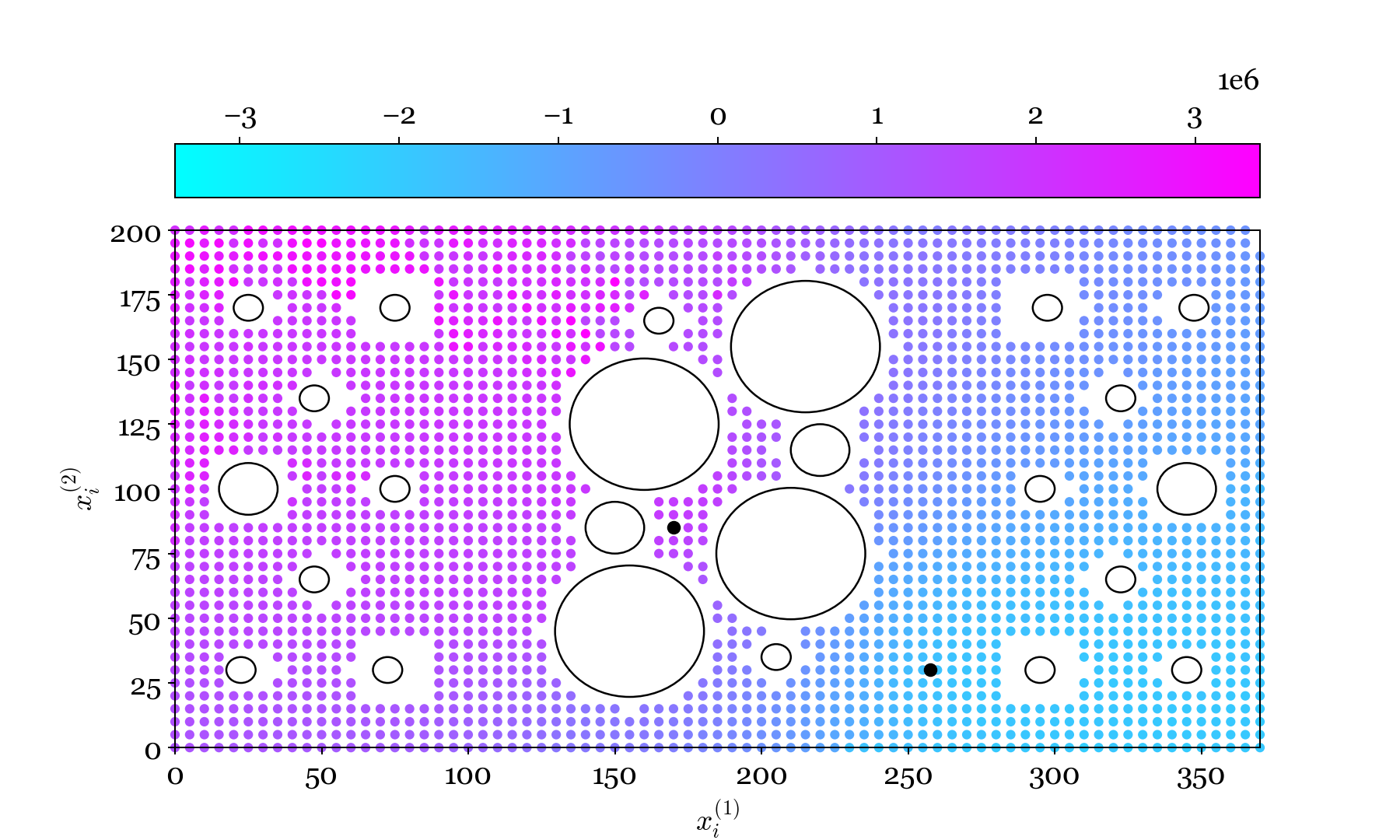}
\caption{Left: all source locations (blue $\bullet$) and sensor locations (black $\bullet$). Right: heatmap of the $\Delta$ToA (response $y_i$) given locations (inputs $\mathbf{x}_i = \{x^{(1)}_i, x^{(2)}_i\}$) for experiment $k=15$, sensor pair (3, 5). }
\label{fig:plate-locs}
\end{figure}

Artificial AEs were excited by thermoelastic expansion generated with an incident laser pulse, with the signals captured at 8 piezoceramic (Sonox P5) sensors mounted to the surface of the plate, visible in \Cref{fig:real-plate} and shown by the black markers in \Cref{fig:plate-locs}, numbered 1-8 (left). %
The sensors operate by converting surface displacements resulting from the AE stress waves into electrical energy, allowing the ultrasonic signals to be captured digitally. %
There are a total of $N=2227$ possible source locations, shown by blue markers in \Cref{fig:plate-locs} (left). %
The time of arrival for each of the 8 sensors was extracted following standard practice, using an autoregressive form of the Akaike Information Criterion (AIC)~\cite{hensman2010locating}. %

\paragraph{Difference in time-of-arrival ($\Delta$ToA)} Let the arrival time of AE $i$ at sensor $j$ be denoted $A_{ij} \quad \forall j \in \{1, 2, \ldots, 8\}$. %
The difference in time-of-arrival ($\Delta$ToA) is then the difference between any two sensors. %
Since there are 8 sensors, there are 28 pairwise combinations and associated maps (8C2). Each pair generates a different $f$ (distinguished with notation later). %
For example, the pair (3, 5) would present the scalar output $y_i \in \mathbb{R}$,
$$ y_i = A_{i3} - A_{i5}$$

\begin{figure}[ht]
\centering
\includegraphics[width=\linewidth]{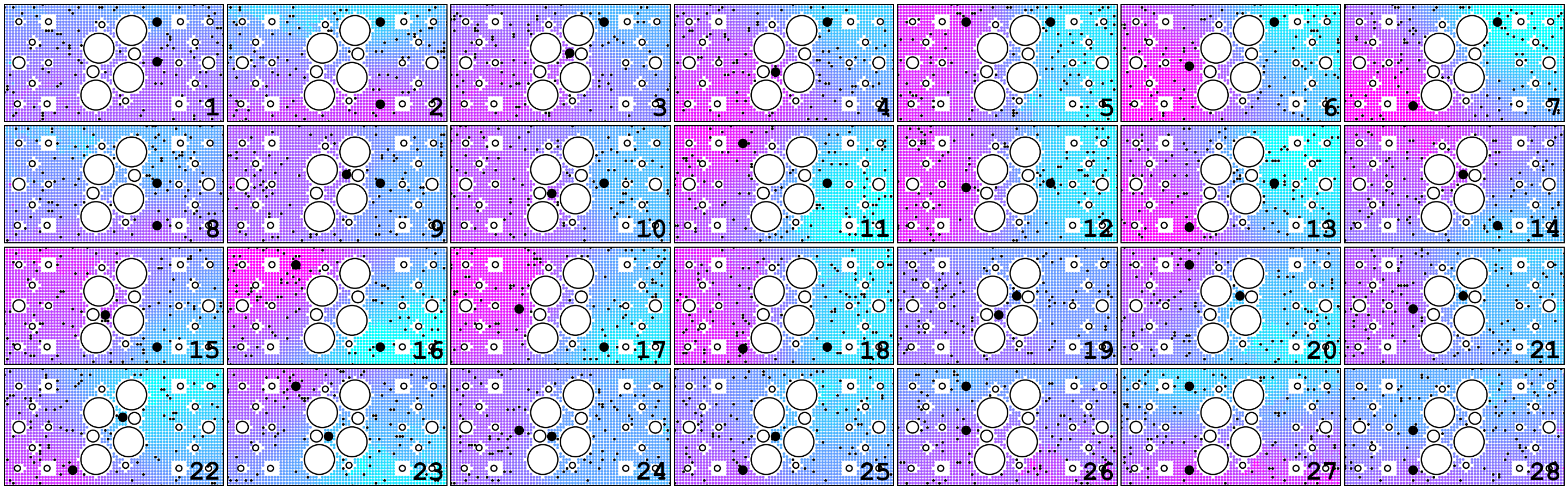}%
\caption{Heatmaps of the measured $\Delta$ToA ($y_i$) with respect to locations ($\mathbf{x}_i$) for all sensor combinations. Large black markers plot sensor pairs, while small black markers plot training observations. The remaining data are used to test out-of-sample performance. The number in the bottom right is the experiment index $k \in \{1,2, ...28\}$.}
\label{fig:allpairs}
\end{figure}

\noindent The input vectors $\mathbf{x}_i \in \mathbb{R}^2$ are the locations (length vs.\ width), 

$$
\mathbf{x}_i = \{x^{(1)}_i, x^{(2)}_i\}
$$

\noindent where \{0, 0\} is the bottom left corner of the plate. %
\Cref{fig:plate-locs} (right) shows the data associated with the sensor pair (3, 5) while \Cref{fig:allpairs} shows all pairs. %
Each combination is labelled with an experiment index $k \in \{1,2...,28\}$ listed in \Cref{t2}, \Cref{a:pair_labels}. %

\paragraph{A note on normalisation}
\Cref{fig:allpairs} plots $N=100$ training data for each sensor pair combination, sampled uniformly from each task. %
We then normalise inputs with respect to the longest edge, such that the plate's length is unity. %
This scaling maintains approximately the same relative smoothness of the map in each direction, i.e. one length scale for both dimensions.
Rather than normalise the response for each plate independently, it is z-score normalised with respect to all 28 sensor pairings. %
This maintains the relative structure between experiments (sensor pairs) in the combined data. %
More specifically, a \textit{global} normalisation of the outputs is essential to prevent meaningful differences (between tests) from being scaled out of the data. %

\paragraph{Why multilevel?}

In the context of this work, each map and sensor pair corresponds to a distinct but related \textit{environment}: each referred to as an \textit{experiment}, with an associated predictive task learnt as a regression. %
These collected environments are visualised in \Cref{fig:allpairs}. %
If one learns these maps collectively, in a combined inference, the model should capture variations between each sensor pair. %
These relationships capture further insights from the experimental campaign, which emerge at the systems level.

\section{Representing AE Maps with Gaussian Processes Regression}\label{s:stl}

GPs are used to represent each of these experiments since they offer a flexible tool for regression with natural mechanisms to encode engineering knowledge and domain expertise. %
First, we consider one task at a time (\Cref{s:mtl} extends the model to consider all experiments in a combined inference). The additive noise $\epsilon_i$ is assumed to be normally distributed,
\begin{equation}
y_i = f(\mathbf{x}_i) + \epsilon_i, \qquad \epsilon_i \sim \textrm{N}(0, \sigma_i)
\end{equation}

\noindent in practice, $\epsilon_i$ will represent more than observation noise, since it captures the combined uncertainty of the $\Delta$ToA extraction, from the raw time series data. %
Still, as with previous work~\cite{jones2022bayesian,jones2022heteroscedastic, jones2023constraining,hensman2010locating} we found this representation works well in practice. 

Rather than suggest a parametrisation of $f$ we assume that it has a nonparametric GP prior distribution\footnote{Note, we use $p(\textbf{a} \mid \textbf{b})$ and $\textbf{a} \sim p(\textbf{b})$ to denote the same probability distribution.},
\begin{align}
    \boldsymbol{\theta} &\sim p(\boldsymbol{\phi}) \label{e:th_samp}\\
    f &\sim \textrm{GP}\left(m_f(\;\cdot\;; \boldsymbol{\theta}),\; k_f(\;\cdot\;,\;\cdot\;; \boldsymbol{\theta}) \right)
\end{align}

The prior over $f$ is specified by its mean $m_f$ and covariance $k_f$ functions, where $\boldsymbol{\theta}$ is the set of collected hyperparameters. %
These hyperparameters are sampled from the higher level distribution $p(\boldsymbol{\theta}_k \mid \boldsymbol{\phi})$ whose \textit{prior} is parametrised by constants in $\boldsymbol{\phi}$. %
The mean and covariance of the prior offer natural mechanisms to encode knowledge of the expected functions given domain expertise, before any data are observed. %
The covariance determines the expected correlation between outputs -- influencing process variance, and smoothness -- while the mean represents the prior expectation of the structure of the functions. %

A function sample from the GP, denoted $\mathbf{f}$, is multivariate normal for any finite set of $N$ observations,

\begin{align}
    \mathbf{f} &\sim \textrm{N}\left(\mathbf{m_f},\; \mathbf{K_f} \right) \\
    \textbf{y} &\sim \textrm{N}(\textbf{f}, \boldsymbol{\sigma}) \\[1em]
    \mathbf{K_f}[i,j] &\triangleq k_f(\mathbf{x}_i, \mathbf{x}_j ; \boldsymbol{\theta}) \\ 
    \mathbf{m_f}[i] &\triangleq  m_f(\mathbf{x}_i) 
\end{align}

\noindent Since the \textit{observation} model is assumed Gaussian, we can avoid sampling $\mathbf{f}$ entirely by combining the GP kernel $\mathbf{K_f}$ with the additive noise vector $\boldsymbol{\sigma}$, to describe the likelihood function,
\begin{align}
\mathbf{y} &\sim \textrm{N}(\mathbf{m_f}, \mathbf{K_f} + \operatorname{diag}(\boldsymbol{\sigma}^2)) \label{e:lik}
\end{align}

\noindent Following~\cite{jones2022bayesian} we use a zero-mean and Mat\'{e}rn $3/2$ kernel function $k_f$ for the covariance function,
\begin{align}
\mathbf{m_f}[i] &= 0 \\
\mathbf{K_f}[i,j] &= \alpha^2\left(1+\frac{\sqrt{3}\left|\mathbf{x}_i-\mathbf{x}_j\right|}{l}\right) \exp \left(-\frac{\sqrt{3}\left|\mathbf{x}_i-\mathbf{x}_j\right|}{l}\right)\label{eq:Matern32}
\end{align}

\noindent The zero-mean function is justified since the complex plate geometry prevents the specification of a parametrised mean. %
In the absence of this information, a zero mean is usually sufficient practice, since the GP alone is flexible enough to model arbitrary trends, given enough training data. %
Two hyperparameters are introduced via the kernel $k_f$ (\ref{eq:Matern32}) such that $\boldsymbol{\theta} = \{\alpha, l\}$. %
The process variance $\alpha$ encodes the magnitude of function variations around the expected mean. %
The length scale $l$ encodes how much influence a training observation has on its neighbouring inputs, it represents the \textit{smoothness} of the approximating family of functions.

\subsection{Heteroscedastic noise}

In the experiments, the scale of the additive noise $\sigma_i$ is expected to increase at the extremities of the plate~\cite{jones2022heteroscedastic}. %
As such, the magnitude of the noise is \textit{input-dependant}, and it is not statistically identical over the whole input (i.e.\ it is not homoscedastic). %
Such input-dependent noise can be represented with a heteroscedastic regression. %
Heteroscedasticity is implemented with another GP in a combined model, mapping the inputs $\mathbf{x}_i$ to the scale of the additive noise $\sigma_i$. %
This introduces the \textit{noise-process} $r$, where the prior utilises a constant mean function\footnote{A constant GP mean function does not imply the posterior (predictive) distribution of $r$ is constant, only the mean of its prior distribution.}
$m_r(\mathbf{x}_i) = m_r$ and another Mat\'{e}rn $3/2$ kernel function (with its own hyperparameters). %
Therefore, a finite sample from the noise process is distributed as follows,

\begin{align}
\textbf{r} &\sim \textrm{N}\left(\mathbf{m_r}, \mathbf{K_r} \right) \\
\boldsymbol{\sigma} &= \exp(\mathbf{r}) \label{e:r-GP} \\[1em]
\mathbf{m_r}[i] &= m_r \nonumber \\
\mathbf{K_r}[i,j] &= \alpha_r^2\left(1+\frac{\sqrt{3}\left|\mathbf{x}_i-\mathbf{x}_j\right|}{l_r}\right) \exp \left(-\frac{\sqrt{3}\left|\mathbf{x}_i-\mathbf{x}_j\right|}{l_r}\right)
\end{align}

\noindent Samples from $r$ are exponentiated to define $\boldsymbol{\sigma}$ since the noise variance must be strictly positive (note that, untransformed, a GP will map to any real number). %
The exponential transformation requires a constant mean function, otherwise, the prior of the expected noise variance is too large, as a zero mean function would lead to $\exp(0) = 1$. %
The additional hyperparameters from the noise process are now included in the total set $\boldsymbol{\theta}$, 

$$\{m_r, \alpha_r, l_r\} \in \boldsymbol{\theta}$$

\subsection{Prior formulations}

One should encode \textit{prior knowledge} of the AE maps as prior distributions over the higher-level variables $\boldsymbol{\theta} \sim p(\boldsymbol{\phi})$. %
The hyperparameters of the GP are sampled from these distributions, which characterise the expected variation, smoothness, and noise of the response. %
The following $p(\boldsymbol{\theta}_k \mid \boldsymbol{\phi})$ structure is adopted, 
\begin{align}
\boldsymbol{\theta} = \{l, \alpha, &m_r, \alpha_r, l_r\}, \qquad  \boldsymbol{\theta} \sim p(\boldsymbol{\phi})\nonumber \\[1em]
l &\sim \textrm{Gamma}(\textsf{shape}=2, \; \textsf{rate}=1) \label{e:prior_start}\\
\alpha &\sim \textrm{Half-Normal}^+(\textsf{scale}= 1)\\[1em]
m_r &\sim \textrm{N}(\textsf{mean} = -0.9, \; \textsf{scale} =1) \label{e:noise_prior_start}\\
l_r &\sim \textrm{Gamma}(\textsf{shape}=2, \; \textsf{rate}=1) \label{e:prior_lr} \\
\alpha_r &\sim \textrm{Half-Normal}^+(\textsf{scale}= 1) \label{e:prior_end}
\end{align}

\noindent where $\textrm{Gamma}$ and $\textrm{Half-Normal}^+$ are the Gamma and positive Half-Normal probability distribution functions respectively. %
The Half-Normal distribution is centred at zero, with support for positive values only. %
The constants in (\ref{e:prior_start})-(\ref{e:prior_end}) correspond to $\boldsymbol{\phi}$, and they are specified given the combined normalised space as weakly informative prior distributions, explained below. %

Recall that the inputs are normalised between [0, 1] with respect to the longest side. %
We expect a smooth $\Delta$ToA map $(f)$ and noise process $(r)$, so the Gamma distributed length scales $l$ (\ref{e:prior_start}) and $l_r$ (\ref{e:prior_lr}) have their mode at $1 = (\textsf{shape} - 1) / \textsf{rate}$ \cite{bda3}. %
The output is z-score normalised, so we expect the process variance $\alpha$ should be less than unity (shrunk further by outliers). %
This is reflected by a half-normal distribution with a unit scale, with an expected value ${\textrm{E}[\alpha]} = \frac{1 \sqrt{2}}{\sqrt{\pi}} = {0.8}$

\noindent The expection of the noise process $r$ is set to $0.08$, to encode pior belief of a signal-to-noise ratio of 10, in terms of expected scale,
\begin{align}
\frac{\textrm{expected process noise}}{\textrm{expected additive noise}} &= \frac{\textrm{E}[\alpha]}{\textrm{E}[\boldsymbol{\sigma}]} = \frac{\textrm{E}[\alpha]}{\textrm{E}[\exp(r)]} = \frac{\textrm{E}[\alpha]}{\exp(\textrm{E}[\mathbf{m_r}])}\\
&= \frac{{1 \sqrt{2}} \times {\pi}^{-0.5}}{\exp(-2.5)} = \frac{0.8}{0.08}
\end{align}

\noindent A vague prior is defined for $\alpha_r$, which indicates high variance in the noise process~$r$, to reflect weak \textit{a priori} knowledge,
\begin{align}
{\textrm{E}[\alpha_r]} = \frac{2 \sqrt{2}}{\sqrt{\pi}} = {1.6}
\end{align}

\noindent A high scale for the noise process is assumed, since it is known that measurement noise increases dramatically at the extremities of the plate, far from the centroid of sensor pairs~\cite{jones2022heteroscedastic}. %

\subsection{Inference and prediction}

To identify the model and make predictions, one can infer the posterior distribution for latent variables $p(\boldsymbol{\Theta} \mid \mathbf{y})$ by conditioning the joint distribution (which encodes domain expertise via the model and the prior specification) on the training data $\mathbf{y}$. %
We use $\boldsymbol{\Theta}$ to generically collect all (unobservable) latent variables, including functions, parameters, and hyperparameters. %
The joint distribution is written as the product of two densities, referred to as the \textit{likelihood} $p(\mathbf{y} \mid \boldsymbol{\Theta})$ (or the data distribution) and the \textit{prior} $p(\boldsymbol{\Theta})$,
\begin{align}
p(\mathbf{y}, \boldsymbol{\Theta}) = p(\mathbf{y} \mid \boldsymbol{\Theta})p(\boldsymbol{\Theta} \; ; \; \boldsymbol{\phi}) \label{eq:joint}
\end{align}

\noindent During model design, we have specified the likelihood with (\ref{e:lik}) and the prior throughout (\ref{e:prior_start})-(\ref{e:prior_end}).
Applying the property of conditional probability to (\ref{eq:joint}) we arrive at Baye's rule and an expression for the posterior distribution,
\begin{align}
p(\boldsymbol{\Theta} \mid \mathbf{y}) = \frac{p(\mathbf{y} \mid \boldsymbol{\Theta}) p(\boldsymbol{\Theta})}{p(\mathbf{y})} \label{e:bayes}
\end{align}

\noindent While (\ref{e:bayes}) is a straightforward application of conditioning~\cite{murphy2012machine}, in practice, the evaluation of the denominator (i.e. the \textit{marginal likelihood} or \textit{evidence}) is non-trivial. %
It is specified by the following integral, which is intractable for most prior-likelihood combinations,
\begin{align}
\textrm{evidence: \qquad}& p(\mathbf{y}) = \int p(\mathbf{y}, \boldsymbol{\Theta})\; d\boldsymbol{\Theta} = \int p(\mathbf{y} \mid \boldsymbol{\Theta}) p(\boldsymbol{\Theta})\; d\boldsymbol{\Theta} \label{e:mL}
\end{align}

\noindent The integral (\ref{e:mL}) is feasible for a subset of likelihood-prior distributions, known as conjugate pairs~\cite{bda3}. %
In many practical applications, however, it becomes increasingly hard to justify the model and prior choices that lead to conjugacy. %
In our case, the prior formulation $p(\boldsymbol{\theta}\mid \boldsymbol{\phi})$ required for a multilevel representation leads to an intractable~\cref{e:mL}.

A number of approximate Bayesian methods are used with non-conjugate models. %
Here, we utilise a sampling-based solution, inferring the parameters using MCMC and the no U-turn implementation of Hamiltonian Monte Carlo~\cite{hoffman2014no}. %
The models are implemented in the probabilistic programming language \texttt{Stan}~\cite{carpenter2017stan}. %

When predicting new (previously unobserved) data $\mathbf{\tilde{y}}$ one (empirically) integrates out $\boldsymbol{\Theta}$ from the following product,
\begin{align*}
    p(\mathbf{\tilde{y}} \mid \mathbf{y}) &= \int p(\mathbf{\tilde{y}} \mid \boldsymbol{\Theta}) p(\boldsymbol{\Theta} \mid \mathbf{y}) \, d\boldsymbol{\Theta}
\end{align*}

\noindent For GP variables $\{f, r\}$, the distribution used to predict new inputs has an analytical solution when the hyperparameters $\boldsymbol{\theta}$ are fixed. %
For the $f$-process, this would be $p(\mathbf{\tilde{y}} \mid \mathbf{y}, \boldsymbol{\theta}_s)$ where $\boldsymbol{\theta}_s$ represents samples from the approximated posterior distribution. %
The analytical solution is then defined by conditioning a joint Gaussian, e.g.\ $p(\mathbf{\tilde{y}}, \mathbf{y} \mid \boldsymbol{\theta}_s)$, on the training variables for all samples from the approximated posterior distribution. %
The relevant identity is provided in \Cref{a:MI}.

\begin{figure}[ht]
\centering
\includegraphics[width=.55\linewidth, valign=b]{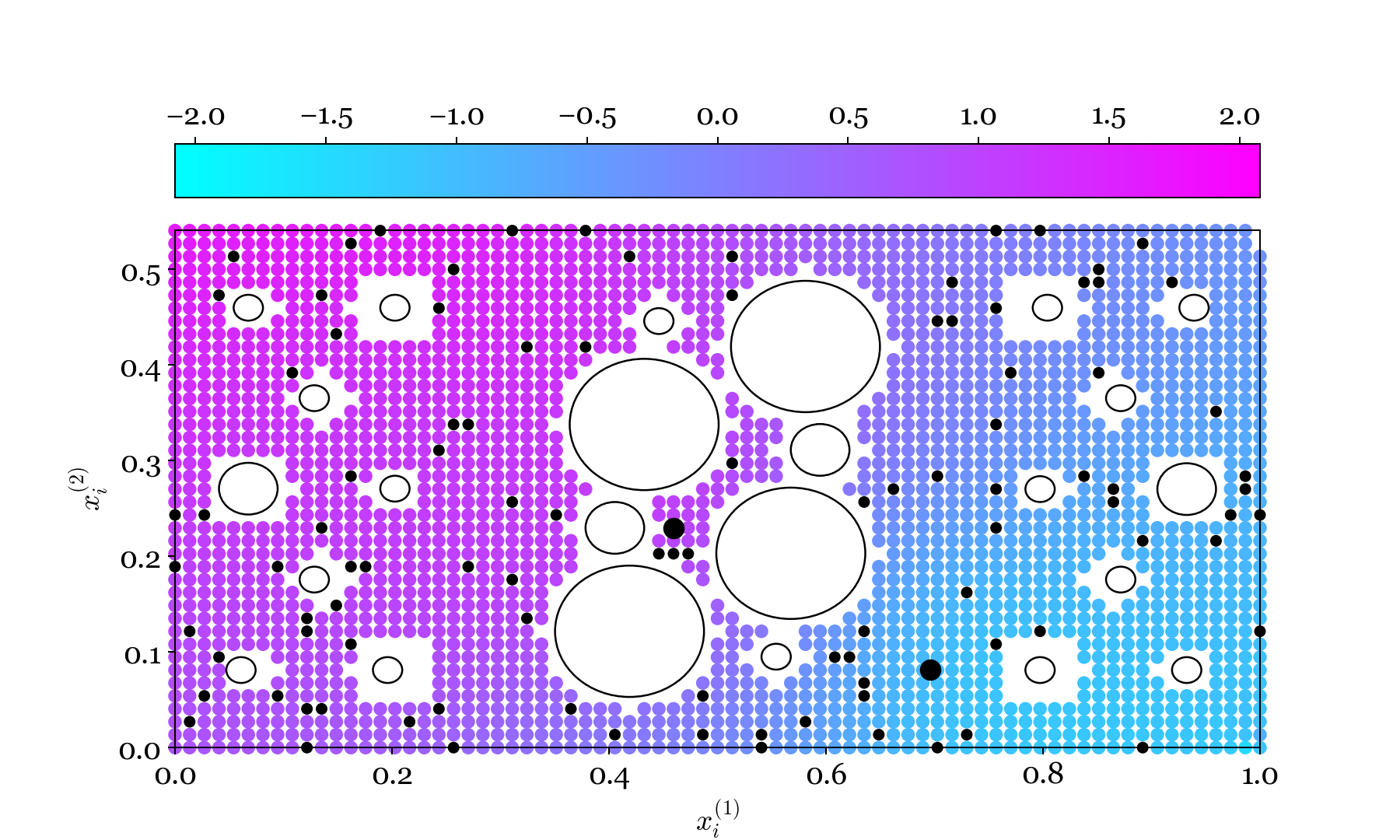}%
\includegraphics[width=.45\linewidth, valign=b]{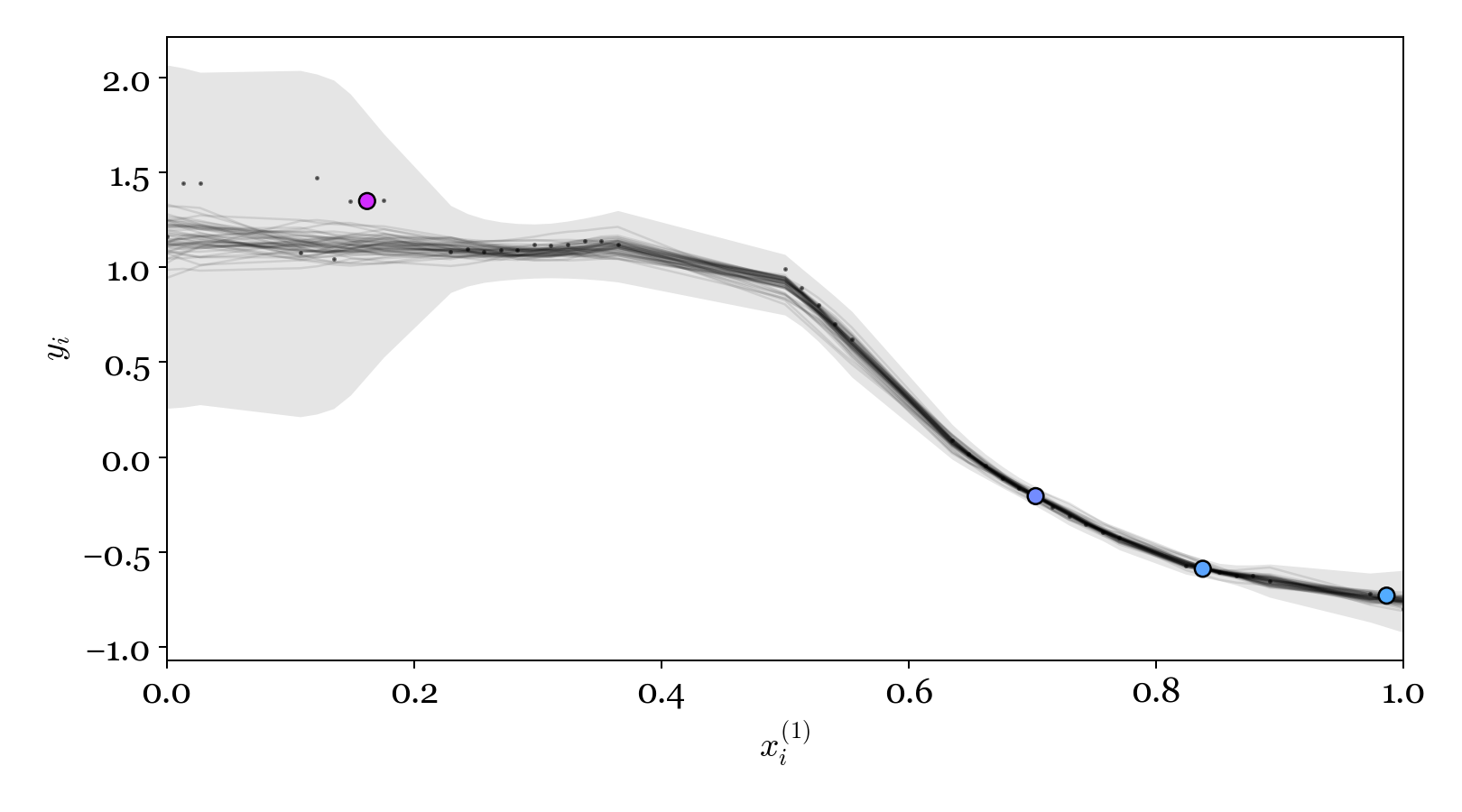}
\caption{The inferred sensor pair (3,5) for experiment $k=15$ (left) and a length-wise slice to visualise the heteroscedastic noise (right).}
\label{f:35map}
\end{figure}

For the (3,5) sensor pair, a random sample of $N=100$ observations is used for training, and the rest are set aside as test data. %
Following inference and prediction, \Cref{f:35map} plots the mean of the posterior predictive distribution for the two-dimensional map over the plate. %
To visualise the predictive variance and its heteroscedastic nature, a random slice is taken along the length of the map and also plotted in \Cref{f:35map}.

\section{Multilevel Representations}\label{s:mtl}

To conceptualise the hierarchical structure of the experimental data, independent GPs are learnt for all 28 tests (rather than sensor pair (3, 5) only). %
Recall that each pairwise map is distinguished with an index $k \in \{1,2,\ldots, 28\}$ listed in \Cref{t2}, \Cref{a:pair_labels}. %
The set of all maps is given by,
$$
\{f_k\}^K_{k=1}
$$

\begin{figure}[ht]
\centering
\includegraphics[width=\linewidth]{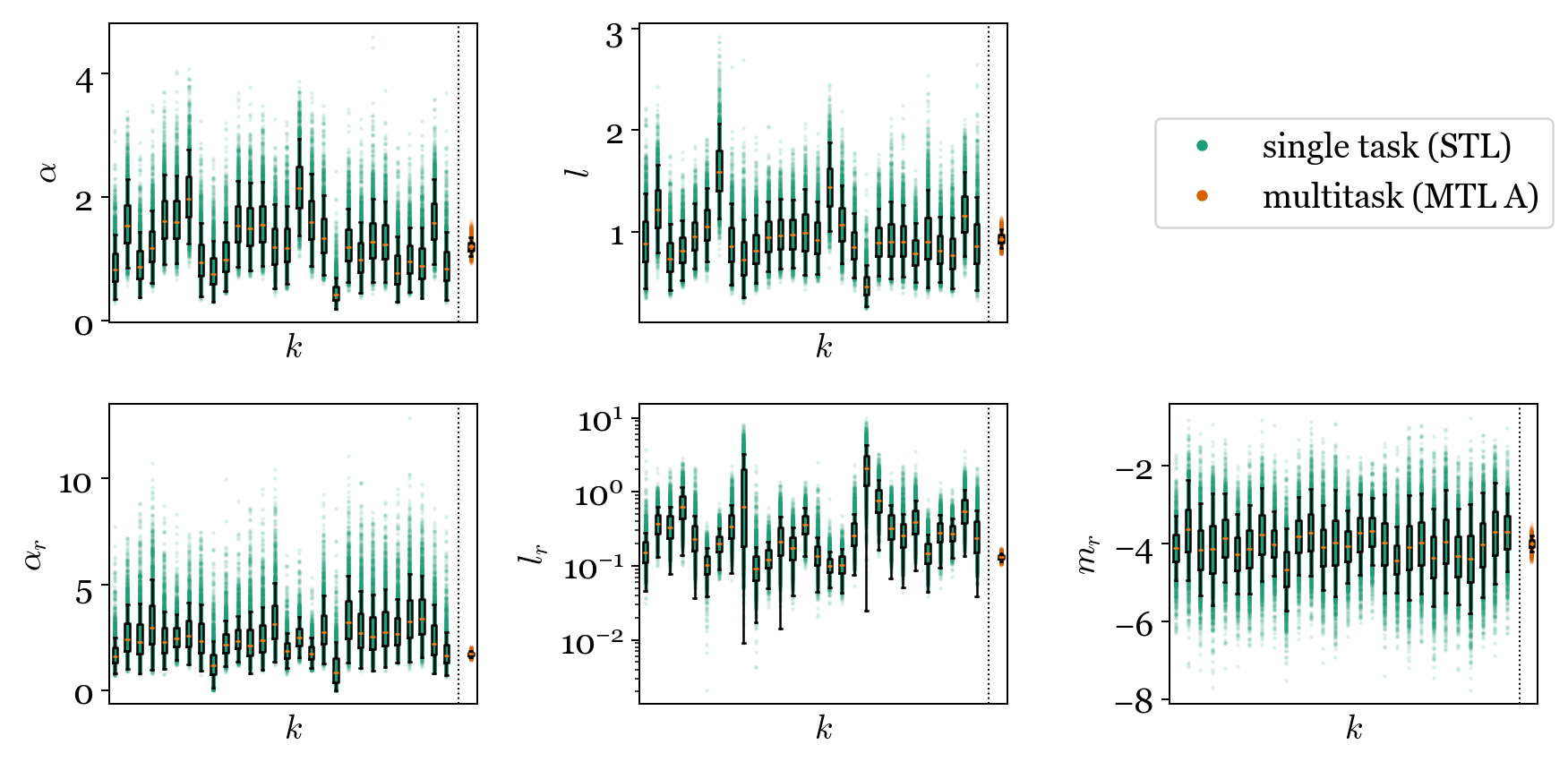}
\caption{Posterior distribution of hyperparameters for independent GPs of each experiment (STL, green), compared to sharing hyperparameters (the prior) between all experiments (MTL-A, orange). Top row: $f$-process hyperparameters (AE map). Bottom row: $r$-process hyperparameters (heteroscedastic noise).}
\label{f:indep_post}
\end{figure}

\noindent Throughout, we use the same test-train split of $N=100$ random samples from each task, visualised in \Cref{fig:allpairs}. %
\Cref{f:indep_post} shows samples from the posterior distributions $p(\boldsymbol{\theta} \mid \mathbf{y}_k) \; \forall k$. %
Since the data are not shared between the experiments, each task is learnt independently, and these models are considered single-task learners (STL)~\cite{murphy2012machine}. %
\Cref{f:indep_post} is typical longitudinal or panel data~\cite{bda3} where the task-specific $\boldsymbol{\theta}$ are similar, with random variations. %
The experiment-specific models can be viewed as perturbations around an average (higher-level model). %
An intuitive concept, since all experiments concern the same plate, despite the varying experimental design (sensor placement). %

To summarise, each task represents a distinct but related environment. %
In turn, MTL is used to learn predictive maps in a combined inference, to capture the inter-task functions (with quantified uncertainty) relating to differences between tests in the experimental campaign.

\subsection{(Method A) K GPs, one prior}

In a combined inference, one simple way to share information between tests assumes that the smoothness $l$ and process variance $\alpha^2$ of all $\Delta$ToA maps $\{f_k\}$ are consistent. %
In other words, the GPs for each experiment are sampled from a shared \textit{prior},
\begin{align*}
\{\mathbf{f}_k\}_{k=1}^K &\sim \textrm{N}\left(\mathbf{m_f},\; \mathbf{K_f} \right) \\
\end{align*}

\noindent Same for the noise process $r_k$,
\begin{align*}
\{\textbf{r}_k\}_{k=1}^K &\sim \textrm{N}\left(\mathbf{m_r}, \mathbf{K_r} \right)
\end{align*}

\noindent Shared hyperparameters across all $K$ GPs is one form of partial pooling~\cite{bda3} and it will likely improve inference compared to independent learners. %
However, \Cref{f:indep_post} suggests the assumption of tied hyperparameters is too simple. %
Furthermore, process understanding supports this concern: it is unlikely that the independent posterior samples correspond to the same hyperparameter since there are differences in the design of each experiment. %
The question is whether the assumption provides a sufficient model. %

Considering the above, distinct maps for each experiment $\{f_k\}$ are sampled from GP priors with shared hyperparameters distributed by $p(\boldsymbol{\theta})$. %
This approach is presented as a benchmark -- multitask learning method A (MTL-A). %
The resultant hyperparameter posterior distributions are shown by orange markers in \Cref{f:indep_post}. %
Their reduced variance is typical for pooled estimates because a single distribution is inferred from all $K$ datasets (rather than one for each task). %
Caution is required, this assumption misrepresents the population variance if the model form is misspecified. %

\subsection{(Method B) Hyperparameter modelling}

A more intuitive generating process considers that certain hyperparameters are conditioned on the experimental setup, rather consistent, to better represent the differences between each test. %
We extend the notation of (\ref{e:th_samp}) with a $k$-index to reflect this,
\begin{align}
\boldsymbol{\theta}_k &\sim p(\boldsymbol{\phi}) \\
\boldsymbol{\theta}_k &\triangleq \{\alpha_k, l_k\}\nonumber
\end{align}
\noindent Distinct variables are considered for the $\mathbf{K_f}$ kernel only ($\alpha_k$ and $l_k$) since these are easier to interpret and relate to domain knowledge, especially if the plate had represented a simple geometry -- this choice also helps to constrain the model design. %
Having a distinct set $\boldsymbol{\theta}_k$ allows the characteristics of the map to vary between each experiment: this should be expected since they are different experimental designs. %
The higher-level sampling distribution $p(\boldsymbol{\theta}_k \mid \boldsymbol{\phi})$ remains shared between all experiments (and leant from pooled data to share information). %
We now consider modifications to the prior model $p(\boldsymbol{\theta}_k \mid \boldsymbol{\phi})$ to encode knowledge and domain expertise of the intertask relationships, i.e.\ between the experiments. %

\paragraph{Beyond exchangeable experiments}
When specifying a new prior model $p(\boldsymbol{\theta}_k \mid \boldsymbol{\phi})$ it is important to consider whether the experiments are exchangeable~\cite{bda3}. %
Here, they would be considered exchangeable if no information is available to distinguish the $\boldsymbol{\theta}_k$'s from one another. %
In other words, the set $\{\boldsymbol{\theta}_k\}_{k=1}^{k}$ (presented \Cref{f:indep_post}) cannot be re-ordered such that patterns are revealed in the latent variables. %

However, since we know about the design of the AE experiments, there are a few possibilities. %
One simple description is sensor separation, to order the models and reveal structure (i.e. patterns) in the hyperparameter estimates. %
Once structure appears, the experiments are no longer exchangeable. %
Sensor separation $\delta S_k$ is defined as the Euclidean distance between any two of the eight $\binom{8}{2}$ possible sensor locations $\mathbf{\hat{S}} = \{\mathbf{\hat{s}}_1, \mathbf{\hat{s}}_2, \ldots \mathbf{\hat{s}}_8\}$,
\begin{align}
\delta S_k = \lvert \mathbf{\hat{s}}_{g} - \mathbf{\hat{s}}_{h} \rvert \quad \forall \quad \textrm{pairs from} \; \mathbf{\hat{S}}
\end{align}
where the vector $\boldsymbol{\delta S}$ will be length $K= \binom{8}{2} =28$. %
The values for sensor separation from each experiment are also presented in \Cref{t2}, \Cref{a:pair_labels}. %
The simple pattern we expect is that process variance $\alpha_k$ will greater for sensors that are further apart; %
i.e. the variation in the $\Delta$ToA  signal will be greater, since the AE signals have the potential to travel further. %

\begin{figure}[ht]
    \centering
    \includegraphics[width=.8\linewidth]{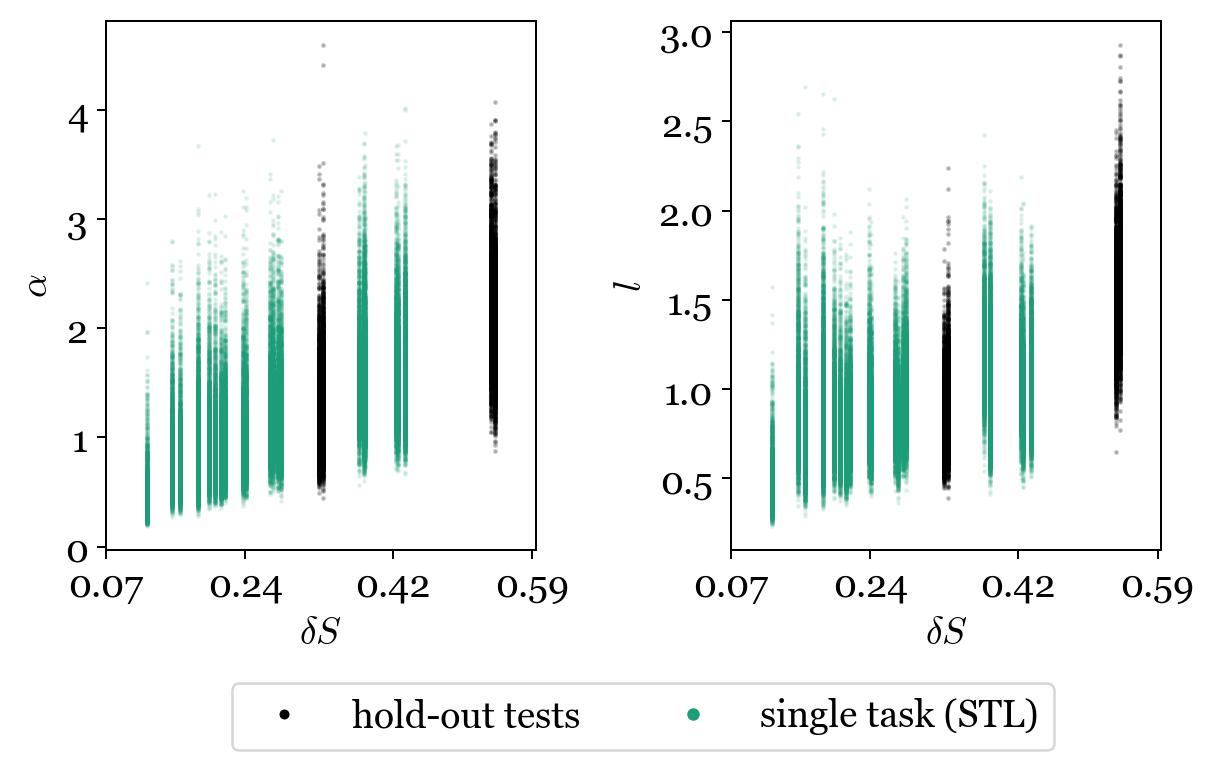}
    \caption{Ordering the hyperparameters (conditional posterior distribution) from the STL experiments $\boldsymbol{\theta}_k$ with respect to sensor separation $\delta S_k$. Black markers correspond models of the hold-out tests $k \in \{4, 7, 16, 22\}$ which do not contribute training data for MTL.}
    \label{f:explained}
\end{figure}

\Cref{f:explained} plots the posterior distributions $\boldsymbol{\theta}_k$ (STL) with respect to $\delta S_k$ for the independent models from \Cref{f:indep_post}. %
These samples indicate parameter relationships that we hope to learn with the higher-level model $p(\boldsymbol{\theta}_k \mid \boldsymbol{\phi})$. %
(Note that in \Cref{f:explained}, samples correspond to latent variables and not observations.)

As expected, the process variance $\alpha_k$ increases with sensor separation. %
The length scale $l_k$ presents a similar trend: we believe the increasing length scale appears for smaller sensor spacing as the noise floor (observation noise as well as parameter uncertainty) is larger compared to the magnitude of the response variations ($\alpha_k$). %
In turn, the influence of each training observation on its neighbours is reduced. %

\Cref{f:explained} also highlights the hold-out experiments $k \in \{4, 7, 16, 22\}$ represented by black markers (rather than green). %
Herein, hold-out experiments are not used when training (as with held-out data) to test interpolation and extrapolation in the hyperparameter space for the multilevel models. %

\paragraph{Intertask GPs}
Like the input-dependent noise process ($r$) the variation of the hyperparameters (of $\mathbf{K_f}$) can be represented with another function. %
The output of these functions is \textit{task-dependent} (i.e.\ experiment-dependent). %
That is, the hyperparameters of AE map $f$ are sampled from higher-level functions, denoted $\{g, h\}$, %
\begin{align}
    \boldsymbol{\theta} &\sim p(\boldsymbol{\phi})\\
    (\alpha) \quad \textrm{g-process}: \quad g &\sim \textrm{GP}\left(m_g(\;\cdot\; ; \boldsymbol{\theta}),\; k_g(\;\cdot\;,\;\cdot\; ; \boldsymbol{\theta}) \right) \\
    (l) \quad \textrm{h-process}: \quad h &\sim \textrm{GP}\left(m_h(\;\cdot\; ; \boldsymbol{\theta}),\; k_h(\;\cdot\; ,\;\cdot\; ; \boldsymbol{\theta}) \right)
\end{align}

\noindent where hyperparameter functions vary (between tasks) with respect to sensor separation $\delta S_k$ for the 24 training experiments,
\begin{align}
\textbf{g} &\sim \textrm{N}\left(\mathbf{m_g}, \mathbf{K_g} \right) \\
\alpha_k &= \textrm{softplus}(g_k) \qquad \forall k \in \textrm{train} \label{e:hg-gp1} \\[1em]
\textbf{h} &\sim \textrm{N}\left(\mathbf{m_h}, \mathbf{K_h} \right) \\
l_k &= \textrm{softplus}(h_k) \qquad \forall k \in \textrm{train}  \label{e:hg-gp2} \\[1em]
\boldsymbol{\theta}_k &\triangleq \{\alpha_k, l_k\}\nonumber
\end{align}

\noindent Each GP is transformed to be strictly positive, as the hyperparameters they predict must be greater than zero. %
While the exponential function was used for this purpose with the noise-process $r$ (\ref{e:r-GP}) here softplus transformation~\cite{wiemann2021using} is used since it can be specified to approximate the identity function ($\,f(a) = a \,$) over the inputs of interest, aiding the interpretability of the hyperparameter models $\{g, h\}$. %
The exponentiated transformation is maintained for the lower-level GPs $f_k$ since it prevented divergences during inference via HMC, and aided convergence of the MCMC chains. %

\paragraph{Priors} The noise process prior distributions are set as before (\ref{e:noise_prior_start})--(\ref{e:prior_end}) while the AE map priors are sampled from GPs (\ref{e:hg-gp1})--(\ref{e:hg-gp2}) whose hyperparameters are added to $\boldsymbol{\theta}$.
We use the same Mat\'{e}rn 3/2 kernel for the higher-level GPs ($k_g$ and $k_h$), however, we use a linear mean function (with slope $\beta$ and gradient $\gamma$), %
\begin{align}
\mathbf{K_g}[k,l] &\triangleq k_g(\delta S_k,\delta S_l; \boldsymbol{\theta})  \nonumber \\
\mathbf{K_h}[k,l] &\triangleq k_h(\delta S_k,\delta S_l; \boldsymbol{\theta})  \nonumber \\
\mathbf{m_g}[k] &\triangleq  \beta_g\delta S_k + \gamma_g \nonumber \\
\mathbf{m_h}[k] &\triangleq  \beta_h\delta S_k + \gamma_h \nonumber 
\end{align}

The linear mean allows us to encode domain knowledge of a positive gradient (as a soft constraint) for the expected intertask functions via the GP prior. %
The conditional posterior predictive distributions of $\{g, h\}$ then provide intertask relationships learnt from the collected data (while only observing data as inputs on the lowest level $f$). %
The additional hyperparameters for the shared (global) sampling distributions are,
$$
\{\alpha_g, l_g, \alpha_h, l_h, \beta_g, \gamma_g, \beta_h, \gamma_h\} \in \boldsymbol{\theta}
$$
Where $\beta$ and $\gamma$ are the slope and intercepts of the linear mean function for the intertask relationships. %
The priors for each of these should reflect the difficulty in making specific statements around hyperparameter values, due to their limited interpretability, %
\begin{align}
    \{ \alpha_g, l_g, \alpha_h, l_h\}& \overset{\textrm{i.i.d.}}{\sim} \textrm{Gamma}(2,1) \label{e:prior_Kfg} \\
    \{\beta_g, \beta_h\} \overset{\textrm{i.i.d.}}{\sim} \textrm{Uniform}(0, 10),& \qquad\{\gamma_g, \gamma_h\} \overset{\textrm{i.i.d.}}{\sim} \textrm{Uniform}(-1, 1)
\end{align}
Given the normalised space, these priors encode weak knowledge and constrain the mean function to positive gradients. %
The upper bound (10) of the uniform distribution was set to avoid divergences of the HMC samples and ensure convergence of MCMC. %
The descriptive multilevel model, with GPs representing hyperparameter variations, is referred to as multitask learning method B (MTL-B). %

\subsection{MTL comparison}

\Cref{table:methods} is provided to compare the parameter hierarchies between each method:
\begin{itemize}
    \item Independent learners (STL)
    \item Shared GP prior (MTL-A)
    \item Hyperparameter modelling (MTL-B)
\end{itemize}   %

STL has distinct GPs $\{f, r\}$ and associated hyperparameters $\{\alpha, l, m_r, \alpha_r, l_r\}$ for each task. %
MTL-A has distinct task GPs $\{f, r\}$ but shares hyperparameters $\{\alpha, l, m_r, \alpha_r, l_r\}$ between all tasks.
Lastly, MTL-B has distinct GPs $\{f, r\}$ and hyperparameters of the response $\{\alpha, l\}$ which are themselves predicted by intertask GPs $\{g, h\}$, with (shared) hyperparameters $\{\alpha_g, l_g, \alpha_h, l_h\}$.

\begin{table}[h]
\centering
\begin{tabular}{|l||c|c|c|}
\hline
\textbf{Method} & \textbf{STL} & \textbf{MTL-A} & \textbf{MTL-B} \\ \hline \hline
\textbf{Task-specific} & 
\begin{tabular}[c]{@{}c@{}} $\{f, r\}$ \\ $\{\alpha, l\}$ \\ $\{m_r, \alpha_r, l_r\}$ \end{tabular} & 
$\{f, r\}$ & 
\begin{tabular}[c]{@{}c@{}} $\{f, r\}$ \\ $\{\alpha, l\}$ \end{tabular} \\ \hline
\textbf{Shared} & 
N/A & 
\begin{tabular}[c]{@{}c@{}} $\{\alpha, l\}$ \\ $\{m_r, \alpha_r, l_r\}$ \end{tabular} & 
\begin{tabular}[c]{@{}c@{}} $\{g, h\}$ \\ $\{\alpha_g, l_g\}$ \\ $\{\alpha_h, l_h\}$ \\ $\{m_r, \alpha_r, l_r\}$ \end{tabular} \\ \hline
\end{tabular}
\caption{Comparison of methods based on the hierarchy of latent variables}
\label{table:methods}
\end{table}

\begin{figure}[ht]
    \centering
    \includegraphics[width=\linewidth]{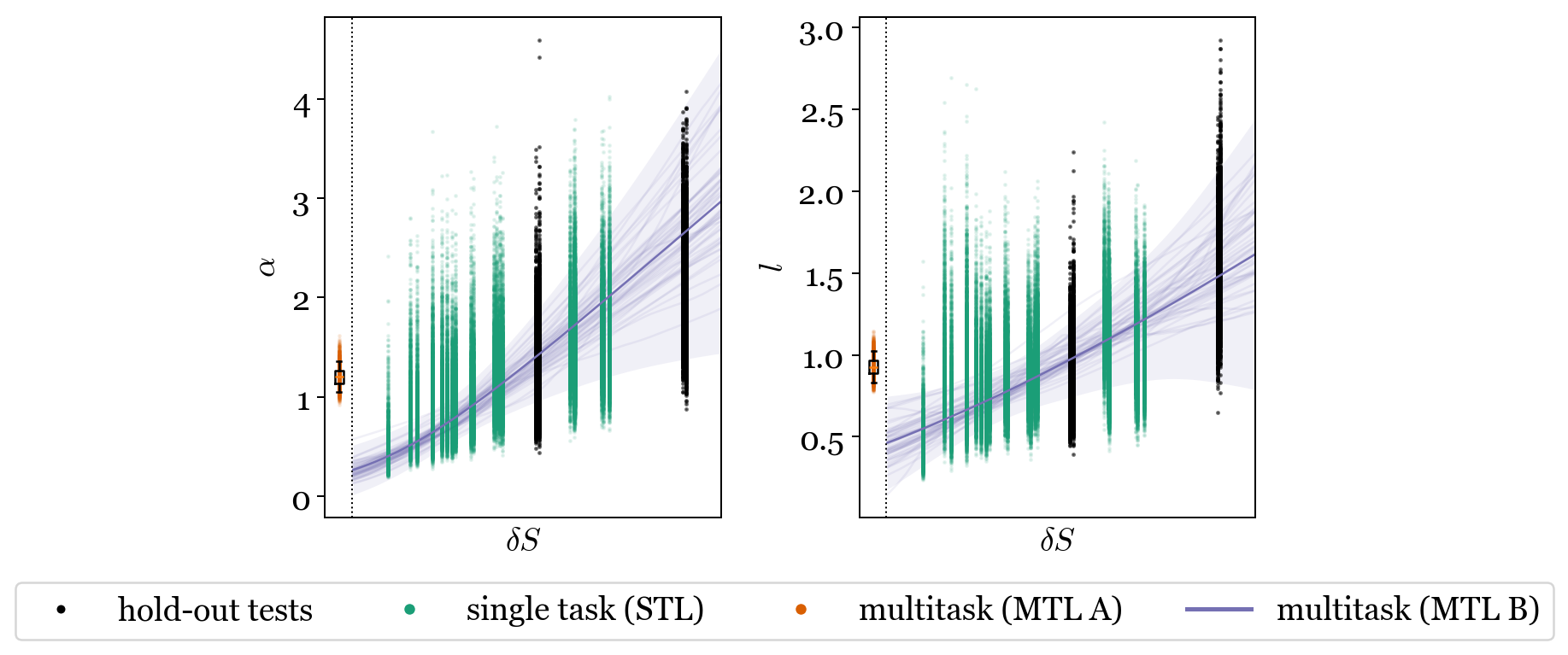}
    \caption{Posterior distributions $p(\boldsymbol{\theta}_k\mid \mathbf{y})$ following each method of learning from the series of AE experiments. Independent learners (STL, green); Shared GP prior (MTL-A, orange); Hyperparameter modelling (MTL-B, purple). Black markers correspond to STL models of the hold-out tests $k \in \{4, 7, 16, 22\}$ which do not contribute training data for MTL.}
    \label{fig:all_models}
\end{figure}

For visual comparison, \Cref{fig:all_models} plots the resultant hyperparameter posterior distributions for STL (green), MTL-A (orange), and MTL-B (purple). %
The (purple) functions $\{g, h\}$ appear to capture the hyperparameter relationships, compared to the trends presented by the posterior distributions of the STL models. %
The variance of the functions is also reduced, compared to independent models (green), without the assumption of one consistent hyperparameter (orange). %
Critically, while each hyperparameter model makes sense given their respective assumptions, the GP model (purple) is more expressive: it represents the variations of the map between experiments with respect to an explanatory variable (sensor separation). %

To summarise, both the process variance $\alpha_k$ and length scale $l_k$ increase with sensor separation $\delta S$. %
For process variance $\alpha_k$, this is because variation in the difference in time of arrival will be greater for sensors that are further apart. %
For the length scale $l_k$ we believe lower amplitude signals have a lower signal-to-noise ratio, therefore the inferred functions are less smooth (i.e.\ lower length scale). %

\paragraph{Modelling hyperparameters, not parameters}

Why not include $\delta S_k$ as an explanatory variable on the lower level -- within a product kernel, for example? %
This is because changes in the response are not smooth with respect to $\delta S_k$. %
Instead, the response surface switches discontinuously between experiments, see \Cref{fig:allpairs}. %
However, the response \textit{characteristics} (smoothness, process variance) show smooth relationships with respect to $\delta S_k$ (observed in \Cref{fig:all_models,f:explained}). %
For this reason, it is more appropriate to model variations at the hyperparameter level, rather than the map itself. %
The expected smoothness of intertask functions should be a primary consideration when building multilevel models. %

\paragraph{Post-selection inference} By investigating (independent) model behaviour with plots of the posterior distributions, we use the training data twice: (i) for inference, (ii) to inform model design. %
Specifically, \Cref{f:explained} was used to inform the structure of the multilevel model, so we are guilty of post-selection inference~\cite{lee2016exact}. %
With reference to \citet{bda3}, when the number of candidate models is small, the bias resulting from data reuse is also small. %
In this case study, the plots informed only certain aspects of the higher-level GP design. %
Post-selection inference should be treated cautiously, however, as when the number of candidate models grows, the risk of overfitting the data increases. %

\section{Results and Discussion}\label{s:res}

To test each method of representing the experimental data, the ground truth (or target) out-of-sample data $\{\mathbf{\bar{y}}_k\}_{k=1}^k$ are compared to the posterior predictive distribution. %
The predictive log-likelihood is used as a probabilistic assessment of performance,

\begin{align}
\textrm{lpY}_k =\sum_{i=1}^N \log \left(\frac{1}{S} \sum_{s=1}^S p \big(\bar{y}_{ik} \mid \boldsymbol{\Theta}_s \big) \right) \label{e:lpY}
\end{align}

\noindent where $\{\boldsymbol{\Theta}_s\}_{s=1}^S$ are the $S$ samples from the full approximated posterior distribution and the combined likelihood of all models is $\textrm{lpY} = \sum_{k=1}^K \textrm{lpY}_k$. %
In words, (\ref{e:lpY}) quantifies the (log) likelihood that test data were generated by the model inferred from the training data. %
A higher value indicates that the model has a better approximation of the underlying data-generating process and indicates good generalisation. %

\begin{figure}[ht]
    \centering
    \includegraphics[width=.8\linewidth]{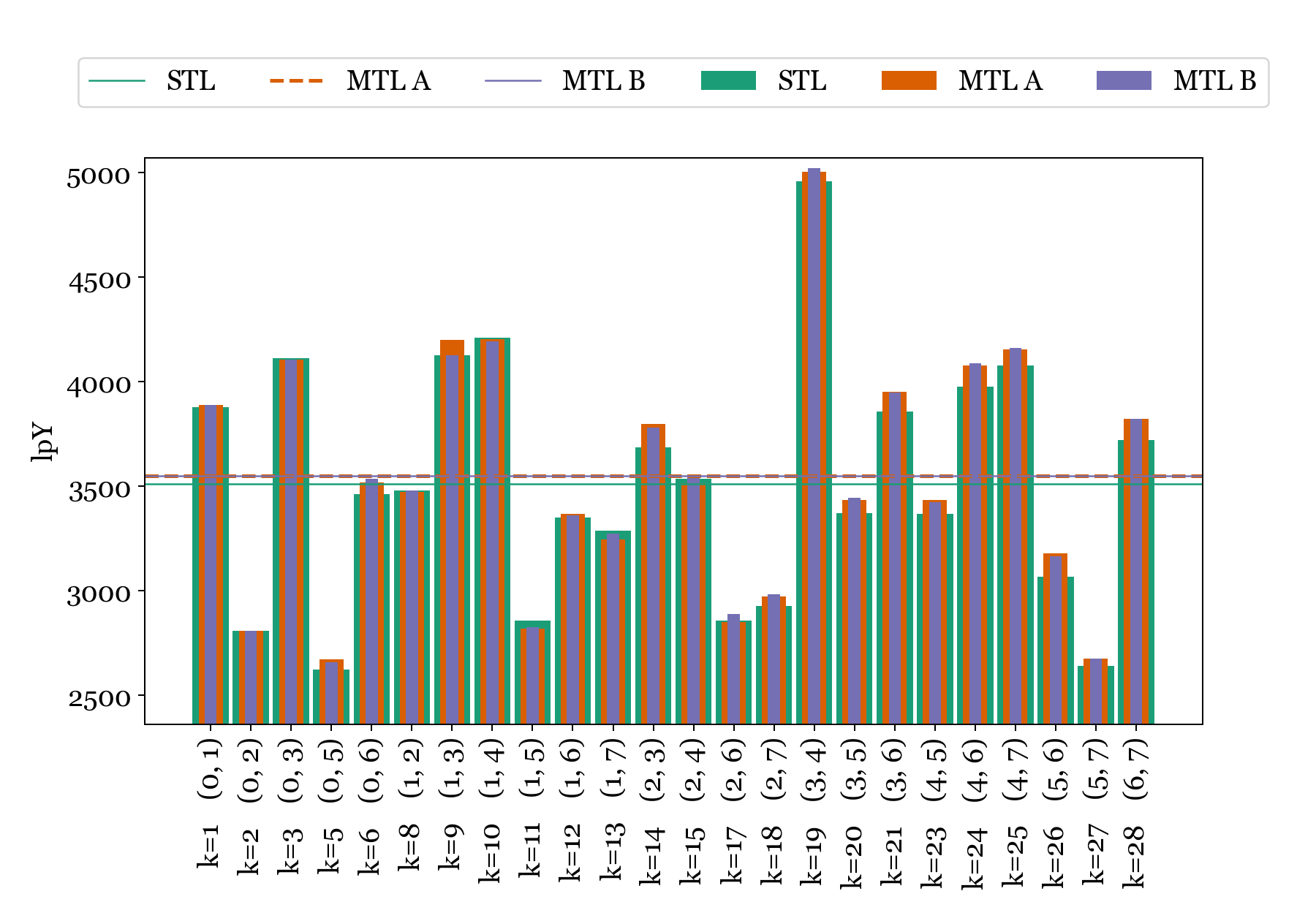}
    \caption{Predictive log-likelihood for the out-of-sample test data. Excluding hold-out experiments $k \in \{4, 7, 16, 22\}$ which are modelled using transfer learning in \Cref{s:transfer}. Lines plot the average across all tasks, the combined log-likelihood (summed) for each method is: (STL, 84217), (MTL~A,~85149), (MTL~B,~85204).}
    \label{fig:logL}
\end{figure}

\Cref{fig:logL} presents $\textrm{lpY}$ for all benchmarks.
There is a small improvement in the combined predictive likelihood for both MTL methods. %
Rather than motivate MTL, the predictive likelihoods show that hierarchical GPs maintain the predictive performance of the STL models. %
Insights are enabled in \Cref{fig:all_models}, where the inferred intertask functions are presented with uncertainty quantification -- e.g.\ these are used later for transfer learning in \Cref{s:transfer}. %
For task-specific performance ($\textrm{lpY}_k$) independent STL performs better for a subset of tasks $k \in \{3, 8, 10, 11, 13\}$. %
This is typical, however, since MTL considers the joint distribution of the whole data. %
If some experiments have data with fewer outliers and less noise, these might be negatively impacted by experiments with higher noise and sparse data. %
For example, in tasks with low sensor separation, the signal-to-noise ratio is lower, which might represent \textit{weak} data. %

Both MTL-A and B extend data for training by partial pooling, but B has a more descriptive multilevel structure, allowing us to encode domain expertise for intertask (between experiment) learning. %
In turn, hyperparameter relationships are captured over the experimental campaign (rather than their marginal distribution). %
By modelling prior variations in MTL-B, it represents changes in the tasks with respect to experimental design parameters (in this case $\delta S_k$).
To summarise:

\begin{itemize}
    \item (A \& B) can simulate the hyperparameters for hypothetical/unobserved experimental setups
    \item (B) has the potential to encode physics/domain expertise of behaviour between sensor pairs (mean functions, constraints)
    \item  (A \& B) can use the intertask relationships, learnt from similar experiments, to predict hyperparameters for tasks with sparse data (this can be viewed as a form of transfer learning)
\end{itemize}

\subsection{Using the multilevel model for transfer learning}\label{s:transfer}
To demonstrate transfer learning, the intertask functions are used for meta-modelling, to interpolate and extrapolate in the model space. %
The results intend to show how multilevel models can capture overarching insights from the experimental campaign (at the systems level) as well as task-specific insights. %
The expected hyperparameter values $\textrm{E}[\boldsymbol{\theta}_k]$ from MTL-A and B can be visualised in \Cref{fig:all_models} with the solid lines. %
These point estimates are used when conditioning on data from the hold-out tests\footnote{Ideally, the full predictive distribution should be used, rather than the expectation alone. A point estimate is used here, however, for computational reasons -- there are 100 repeats of these experiments.}. %
In turn, hyperparameter inference can be avoided for new (previously unobserved) experiments. %
Instead, we predict their value given the other, similar experiments: %
where MTL-A assumes one hyperparameter set for all tests, and MTL-B learns how these vary with respect to sensor separation. %
By predicting hyperparameters, informed by data-rich tasks, the predictive performance should be improved for new experiments with sparse data.

\begin{figure}[ht]
    \centering
    \includegraphics[width=\linewidth]{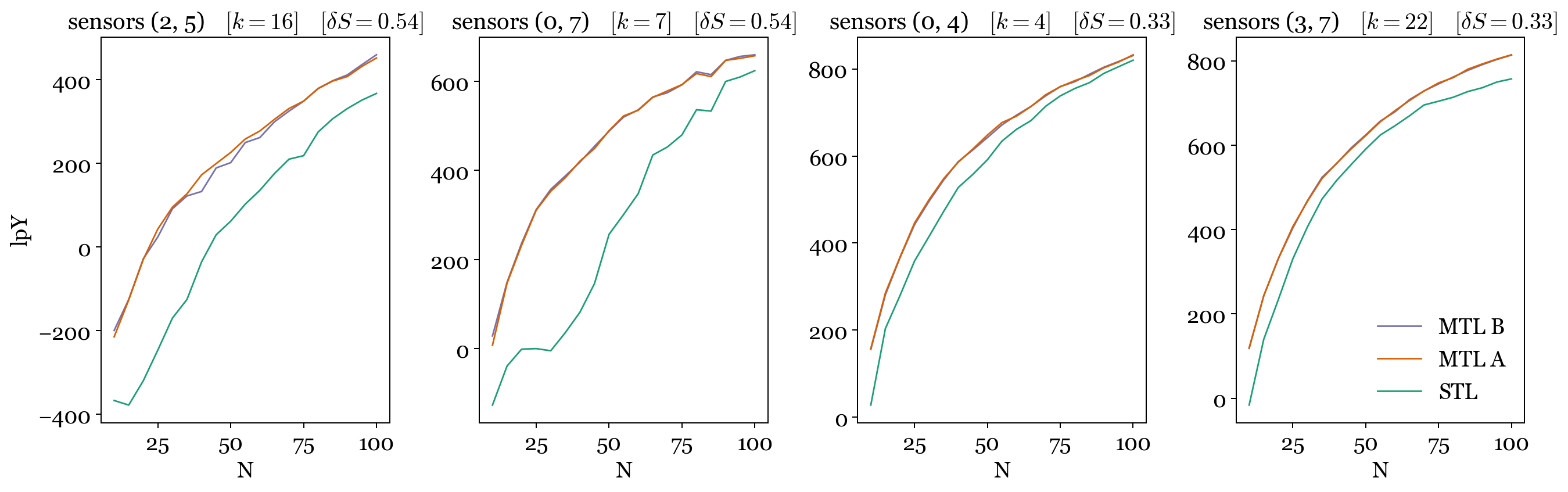}
    \caption{Predictive log-likelihood for an increasing training budget for the hold-out experiments $k \in \{4, 7, 16, 22\}$. Conventional single-task learning (STL) compared to multitask learning (MTL) which predicts hyperparameter values from similar experiments. Averaged for 100 repeats.}
    \label{fig:Ncurves}
\end{figure}

The expected hyperparameter values are used to condition new GPs for an increasing training budget ($N=5 \;\textrm{-}\;100$) for the hold-out experiments. %
\Cref{fig:Ncurves} shows that both forms of MTL consistently improve the predictive performance, especially when extrapolating in the model space ($\delta S = 0.54$). These improvements are intuitive since the extrapolated parameters are associated with higher uncertainty for conventional STL (refer to \Cref{fig:all_models}). %
Another (more natural) way to share information would be to retrain the MTL models and include held-out experiments while increasing the training data. %
We favour the method presented here for computational reasons, referring again to the footnote$^4$. %

These results demonstrate transfer learning since the information from the training experiments (source tasks) is used to improve prediction for held-out experiments (target tasks). %
However, both MTL-A and B provide (effectively) the same performance increase. %
This raises the question: is the more descriptive model for $p(\boldsymbol{\theta} \mid \phi)$ worth it, given that A provides the same predictive performance? %
We argue that this depends on the purpose of the model. %
If the purpose is purely prediction, the assumptions of A will likely be sufficient (in engineering, however, prediction is rarely the only motivation). %
On the other hand, B more closely resembles our understanding and domain expertise of the experimental campaign: it provides more insights regarding variations of the AE map between each experiment. %
In future work, stronger constraints could be applied to the parameters of the model directly (rather than hyperparameters). %
This structure would allow the meta-model to simulate the response $\mathbf{y}_k$ of unobserved experiments directly, and the inter-task functions would have more influence on predictive performance. %
However, these constraints would be more restrictive and require more specific domain knowledge, which was not available for these experiments.

\section{Concluding Remarks}\label{s:conc}

In this work, we demonstrated how multilevel Gaussian Processes (GPs) can be used as meta-models to represent an experimental campaign for non-destructive testing (NDT). %
Two model formulations were used to represent a series of experiments concerning source localisation with acoustic emission (AE) data for a complex plate geometry. %
While the same plate was used throughout the test campaign, the experimental design was varied (re.\ sensor placement). %
By learning all tasks in a joint inference, the representation captures how characteristics of the AE map (i.e.\ hyperparameters) vary between the experiments. %
The model can also share information between tasks, to extend the (effective) number of training data and their value. %
We presented the intertask relations and explained how they inform insights into systems-level behaviour, allowing domain expertise to be encoded between experiments, relating to the effects of design variables on the outcome of each test. %
The intertask functions were used as meta-models to predict hyperparameter values of similar (previously unobserved) experiments and enhance inference in new tasks by transfer learning. %

Looking forward, more specific physics-based constraints could be encoded into multilevel representations (via the kernel function) to describe intertask variations with physics-based models, i.e.\ differential equations. %
Instead, this article encoded general domain expertise of the underlying process (via the GP mean function) such that data simulated from the model reflected our understanding of the environment and experiments.

\section*{Acknowledgements}
LAB and MG acknowledge the support of the UK Engineering and Physical Sciences Research Council (EPSRC) through the ROSEHIPS project (Grant EP/W005816/1). %
AD is supported by Wave 1 of The UKRI Strategic Priorities Fund under the EPSRC Grant EP/T001569/1 and EPSRC Grant EP/W006022/1, particularly the \textit{Ecosystems of Digital Twins} theme within those grants \& The Alan Turing Institute. %
EJC and MRJ are supported by grant reference numbers EP/S001565/1 and EP/R004900/1. %

The support of Keith Worden during the completion of this work is gratefully acknowledged. %
Thanks are offered to James Hensman, Mark Eaton, Robin Mills and Gareth Pierce for their work in generating the data used throughout this paper. %
For the purposes of open access, the authors have applied a Creative Commons Attribution (CC BY) license to any Author Accepted Manuscript version arising.

\bibliographystyle{unsrtnatemph}

\bibliography{ref}

\begin{thebibliography}{27}
\providecommand{\natexlab}[1]{#1}
\providecommand{\url}[1]{\texttt{#1}}
\expandafter\ifx\csname urlstyle\endcsname\relax
  \providecommand{\doi}[1]{doi: #1}\else
  \providecommand{\doi}{doi: \begingroup \urlstyle{rm}\Url}\fi

\bibitem[Gardner \emph{et~al}.(2022)Gardner, Bull, Gosliga, Dervilis, Cross,
  Papatheou, and Worden]{gardner2022pbshm}
P.~Gardner, L.~A. Bull, J.~Gosliga, N.~Dervilis, E.~J. Cross, E.~Papatheou, and
  K.~Worden.
\newblock \emph{Population-Based Structural Health Monitoring}, pages 413--435.
\newblock Springer International Publishing, 2022.

\bibitem[Rahwan \emph{et~al}.(2019)Rahwan, Cebrian, Obradovich, Bongard,
  Bonnefon, Breazeal, Crandall, Christakis, Couzin, Jackson, Jennings, Kamar,
  Kloumann, Larochelle, Lazer, McElreath, Mislove, Parkes, Pentland, Roberts,
  Shariff, Tenenbaum, and Wellman]{Rahwan2019MB}
I.~Rahwan, M.~Cebrian, N.~Obradovich, J.~Bongard, J.-F. Bonnefon, C.~Breazeal,
  J.~W. Crandall, N.~A. Christakis, I.~D. Couzin, M.~O. Jackson, N.~R.
  Jennings, E.~Kamar, I.~M. Kloumann, H.~Larochelle, D.~Lazer, R.~McElreath,
  A.~Mislove, D.~C. Parkes, A.~S. Pentland, M.~E. Roberts, A.~Shariff, J.~B.
  Tenenbaum, and M.~Wellman.
\newblock Machine behaviour.
\newblock \emph{Nature}, 568\penalty0 (7753):\penalty0 477--486, 2019.

\bibitem[Gelman \emph{et~al}.(2013)Gelman, Carlin, Stern, Dunson, Vehtari, and
  Rubin]{bda3}
A.~Gelman, J.~B. Carlin, H.~S. Stern, D.~B. Dunson, A.~Vehtari, and D.~B.
  Rubin.
\newblock \emph{Bayesian Data Analysis}.
\newblock CRC press, third edition, 2013.

\bibitem[Murphy(2012)]{murphy2012machine}
K.~P. Murphy.
\newblock \emph{Machine Learning: A Probabilistic Perspective}.
\newblock MIT press, 2012.

\bibitem[Kreft and De~Leeuw(1998)]{kreft1998introducing}
I.~G. Kreft and J.~De~Leeuw.
\newblock \emph{Introducing Multilevel Modeling}.
\newblock Sage, 1998.

\bibitem[Hensman \emph{et~al}.(2010)Hensman, Mills, Pierce, Worden, and
  Eaton]{hensman2010locating}
J.~Hensman, R.~Mills, S.~Pierce, K.~Worden, and M.~Eaton.
\newblock Locating acoustic emission sources in complex structures using
  {G}aussian processes.
\newblock \emph{Mechanical Systems and Signal Processing}, 24\penalty0
  (1):\penalty0 211--223, 2010.

\bibitem[Karniadakis \emph{et~al}.(2021)Karniadakis, Kevrekidis, Lu,
  Perdikaris, Wang, and Yang]{karniadakis2021physics}
G.~E. Karniadakis, I.~G. Kevrekidis, L.~Lu, P.~Perdikaris, S.~Wang, and
  L.~Yang.
\newblock Physics-informed machine learning.
\newblock \emph{Nature Reviews Physics}, 3\penalty0 (6):\penalty0 422--440,
  2021.

\bibitem[Willard \emph{et~al}.(2020)Willard, Jia, Xu, Steinbach, and
  Kumar]{willard2020integrating}
J.~Willard, X.~Jia, S.~Xu, M.~Steinbach, and V.~Kumar.
\newblock Integrating physics-based modeling with machine learning: {A} survey.
\newblock \emph{arXiv preprint arXiv:2003.04919}, 1\penalty0 (1):\penalty0
  1--34, 2020.

\bibitem[Daw \emph{et~al}.(2021)Daw, Karpatne, Watkins, Read, and
  Kumar]{daw2021physicsguided}
A.~Daw, A.~Karpatne, W.~Watkins, J.~Read, and V.~Kumar.
\newblock Physics-guided neural networks (pgnn): {A}n application in lake
  temperature modeling.
\newblock \emph{arXiv}, 2021.

\bibitem[Wahlstr{\"o}m \emph{et~al}.(2013)Wahlstr{\"o}m, Kok, Sch{\"o}n, and
  Gustafsson]{wahlstrom2013modeling}
N.~Wahlstr{\"o}m, M.~Kok, T.~B. Sch{\"o}n, and F.~Gustafsson.
\newblock Modeling magnetic fields using {G}aussian processes.
\newblock In \emph{2013 IEEE International Conference on Acoustics, Speech and
  Signal Processing}, pages 3522--3526. IEEE, 2013.

\bibitem[Alvarez \emph{et~al}.(2009)Alvarez, Luengo, and
  Lawrence]{alvarez2009latent}
M.~Alvarez, D.~Luengo, and N.~D. Lawrence.
\newblock Latent force models.
\newblock In \emph{Artificial Intelligence and Statistics}, pages 9--16. PMLR,
  2009.

\bibitem[Jones \emph{et~al}.(2023)Jones, Rogers, and
  Cross]{jones2023constraining}
M.~R. Jones, T.~J. Rogers, and E.~J. Cross.
\newblock Constraining {G}aussian processes for physics-informed acoustic
  emission mapping.
\newblock \emph{Mechanical Systems and Signal Processing}, 188:\penalty0
  109984, 2023.

\bibitem[Huang \emph{et~al}.(2019)Huang, Beck, and Li]{huang2019multitask}
Y.~Huang, J.~L. Beck, and H.~Li.
\newblock Multitask sparse {B}ayesian learning with applications in structural
  health monitoring.
\newblock \emph{Computer-Aided Civil and Infrastructure Engineering},
  34\penalty0 (9):\penalty0 732--754, 2019.

\bibitem[Huang and Beck(2015)]{huang2015hierarchical}
Y.~Huang and J.~L. Beck.
\newblock Hierarchical sparse {B}ayesian learning for strucutral health
  monitoring with incomplete modal data.
\newblock \emph{International Journal for Uncertainty Quantification},
  5\penalty0 (2), 2015.

\bibitem[Di~Francesco \emph{et~al}.(2021)Di~Francesco, Chryssanthopoulos,
  Faber, and Bharadwaj]{di2021decision}
D.~Di~Francesco, M.~Chryssanthopoulos, M.~H. Faber, and U.~Bharadwaj.
\newblock Decision-theoretic inspection planning using imperfect and incomplete
  data.
\newblock \emph{Data-Centric Engineering}, 2, 2021.

\bibitem[Papadimas and Dodwell(2021)]{papadimas2021hierarchical}
N.~Papadimas and T.~Dodwell.
\newblock A hierarchical {B}ayesian approach for calibration of stochastic
  material models.
\newblock \emph{Data-Centric Engineering}, 2, 2021.

\bibitem[Hughes \emph{et~al}.(2023)Hughes, Gardner, and
  Worden]{hughes2023towards}
A.~J. Hughes, P.~Gardner, and K.~Worden.
\newblock Towards risk-informed pbshm: Populations as hierarchical systems.
\newblock \emph{arXiv preprint arXiv:2303.13533}, 2023.

\bibitem[Sedehi \emph{et~al}.(2023)Sedehi, Kosikova, Papadimitriou, and
  Katafygiotis]{sedehi2023integration}
O.~Sedehi, A.~M. Kosikova, C.~Papadimitriou, and L.~S. Katafygiotis.
\newblock On the integration of physics-based machine learning with
  hierarchical bayesian modeling techniques.
\newblock \emph{arXiv preprint arXiv:2303.00187}, 2023.

\bibitem[Shull(2002)]{shull2002nondestructive}
P.~J. Shull.
\newblock \emph{Nondestructive evaluation: theory, techniques, and
  applications}.
\newblock CRC press, 2002.

\bibitem[Tobias(1976)]{tobias1976acoustic}
A.~Tobias.
\newblock Acoustic-emission source location in two dimensions by an array of
  three sensors.
\newblock \emph{Non-destructive testing}, 9\penalty0 (1):\penalty0 9--12, 1976.

\bibitem[Jones \emph{et~al}.(2022{\natexlab{a}})Jones, Rogers, Worden, and
  Cross]{jones2022bayesian}
M.~R. Jones, T.~J. Rogers, K.~Worden, and E.~J. Cross.
\newblock A bayesian methodology for localising acoustic emission sources in
  complex structures.
\newblock \emph{Mechanical Systems and Signal Processing}, 163:\penalty0
  108143, 2022{\natexlab{a}}.

\bibitem[Jones \emph{et~al}.(2022{\natexlab{b}})Jones, Rogers, Worden, and
  Cross]{jones2022heteroscedastic}
M.~R. Jones, T.~J. Rogers, K.~Worden, and E.~J. Cross.
\newblock Heteroscedastic {G}aussian processes for localising acoustic
  emission.
\newblock In \emph{Data Science in Engineering, Volume 9: Proceedings of the
  39th IMAC, A Conference and Exposition on Structural Dynamics 2021}, pages
  185--197. Springer, 2022{\natexlab{b}}.

\bibitem[Hoffman \emph{et~al}.(2014)Hoffman, Gelman,
  \emph{et~al}.]{hoffman2014no}
M.~D. Hoffman, A.~Gelman, \emph{et~al}.
\newblock The no-u-turn sampler: adaptively setting path lengths in hamiltonian
  monte carlo.
\newblock \emph{J. Mach. Learn. Res.}, 15\penalty0 (1):\penalty0 1593--1623,
  2014.

\bibitem[Carpenter \emph{et~al}.(2017)Carpenter, Gelman, Hoffman, Lee,
  Goodrich, Betancourt, Brubaker, Guo, Li, and Riddell]{carpenter2017stan}
B.~Carpenter, A.~Gelman, M.~D. Hoffman, D.~Lee, B.~Goodrich, M.~Betancourt,
  M.~Brubaker, J.~Guo, P.~Li, and A.~Riddell.
\newblock Stan: A probabilistic programming language.
\newblock \emph{Journal of statistical software}, 76\penalty0 (1), 2017.

\bibitem[Wiemann \emph{et~al}.(2021)Wiemann, Kneib, and
  Hambuckers]{wiemann2021using}
P.~F. Wiemann, T.~Kneib, and J.~Hambuckers.
\newblock Using the softplus function to construct alternative link functions
  in generalized linear models and beyond.
\newblock \emph{arXiv preprint arXiv:2111.14207}, 2021.

\bibitem[Lee \emph{et~al}.(2016)Lee, Sun, Sun, and Taylor]{lee2016exact}
J.~D. Lee, D.~L. Sun, Y.~Sun, and J.~E. Taylor.
\newblock Exact post-selection inference, with application to the lasso.
\newblock \emph{The Annals of Statistics}, pages 907--927, 2016.

\bibitem[Rasmussen \emph{et~al}.(2006)Rasmussen, Williams,
  \emph{et~al}.]{rasmussen2006gaussian}
C.~E. Rasmussen, C.~K. Williams, \emph{et~al}.
\newblock \emph{Gaussian Processes for Machine Learning}, volume~1.
\newblock Springer, 2006.

\end{thebibliography}

\section*{Appendix}
\appendix
\section{A Gaussian identity for GP prediction}\label{a:MI}
\noindent Let $\mathbf{x}$ and $\mathbf{y}$ be jointly distributed Gaussian random vectors~\cite{rasmussen2006gaussian}, 
$$
\left[\begin{array}{l}
\mathbf{x} \\
\mathbf{y}
\end{array}\right] \sim \textrm{N}\left(\left[\begin{array}{l}
\boldsymbol{\mu}_x \\
\boldsymbol{\mu}_y
\end{array}\right],\left[\begin{array}{ll}
A & C \\
C^{\top} & B
\end{array}\right]\right)=\textrm{N}\left(\left[\begin{array}{l}
\boldsymbol{\mu}_x \\
\boldsymbol{\mu}_y
\end{array}\right],\left[\begin{array}{ll}
\tilde{A} & \tilde{C} \\
\tilde{C}^{\top} & \tilde{B}
\end{array}\right]^{-1}\right)
$$
The 
conditional distribution of $\mathbf{x}$ given $\mathbf{y}$ is
$$ \mathbf{x} \mid \mathbf{y} \sim \textrm{N}\left(\boldsymbol{\mu}_x+C B^{-1}\left(\mathbf{y}-\boldsymbol{\mu}_y\right), A-C B^{-1} C^{\top}\right)
$$
This conditional is used in GP predictive equations, for a given/fixed set (i.e.\ sample) of hyperparameters. %
For further details, refer to \citet{rasmussen2006gaussian}. %

\newpage

\section{Pair label indices}\label{a:pair_labels}
\begin{table}[h!]
\centering
\begin{tabular}{|c|c|c|c|}
\hline
experiment index ($k$) & sensor pair & sensor separation ($\delta S$) \\
\hline
1 & (1, 2) & 0.18 \\
2 & (1, 3) & 0.38 \\
3 & (1, 4) & 0.21 \\
4 & (1, 5) & 0.33 \\
5 & (1, 6) & 0.39 \\
6 & (1, 7) & 0.44 \\
7 & (1, 8) & 0.54 \\
8 & (2, 3) & 0.2 \\
9 & (2, 4) & 0.16 \\
10 & (2, 5) & 0.24 \\
11 & (2, 6) & 0.43 \\
12 & (2, 7) & 0.39 \\
13 & (2, 8) & 0.44 \\
14 & (3, 4) & 0.28 \\
15 & (3, 5) & 0.28 \\
16 & (3, 6) & 0.54 \\
17 & (3, 7) & 0.42 \\
18 & (3, 8) & 0.39 \\
19 & (4, 5) & 0.12 \\
20 & (4, 6) & 0.27 \\
21 & (4, 7) & 0.24 \\
22 & (4, 8) & 0.33 \\
23 & (5, 6) & 0.27 \\
24 & (5, 7) & 0.15 \\
25 & (5, 8) & 0.22 \\
26 & (6, 7) & 0.2 \\
27 & (6, 8) & 0.39 \\
28 & (7, 8) & 0.18 \\
\hline
\end{tabular}
\caption{Experiment indices $k$, sensor pairs, and their separation $\delta S$ (Euclidean distance in normalised space).}\label{t2}
\end{table}

\end{document}